\documentclass[lettersize,journal]{IEEEtran}
\usepackage{amsmath,amsfonts}
\usepackage{algorithmic}
\usepackage{array}
\usepackage[caption=false,font=normalsize,labelfont=sf,textfont=sf]{subfig}
\usepackage{textcomp}
\usepackage{stfloats}
\usepackage{url}
\usepackage{verbatim}
\usepackage{graphicx}
\usepackage{cite}

\usepackage{graphicx}
\usepackage{amsmath}
\usepackage{amssymb}
\usepackage{booktabs}

\usepackage[ruled,linesnumbered]{algorithm2e}
\usepackage{multirow}
\usepackage{array}
\usepackage{graphicx}
\usepackage{verbatimbox}
\usepackage{bm}

\begin{document}

\title{Style Variable and Irrelevant Learning for Generalizable Person Re-identification}

% \author{Haobo Chen*\thanks{Equal Contribution} \quad Chuyang Zhao* \quad Kai Tu \quad Junru Chen \quad Yadong Li  \quad Boxun Li  \\
% MEGVII Technology \\
% {\tt\small \{hbchen121,cy.zhao15\}@gmail.com,
% \{tukai,chenjunru,liyadong,liboxun\}@megvii.com}

\author{Haobo Chen$^*$, Chuyang Zhao$^*$, Kai Tu, Junru Chen, Yadong Li, and Boxun Li
\thanks{Haobo Chen and Chuyang Zhao (\{hbchen121, cy.zhao15\}@gmail.com) have the equal contribution, and this work was done when they were interns at MEGVII Technology.}
\thanks{Kai Tu, Junru Chen, Yadong Li and Boxun Li are are all algorithm research engineers of MEGVII Technology (\{tukai, chenjunru, liyadong, liboxun\}@megvii.com).
Corresponding author: Yadong Li.}
}

% \author{IEEE Publication Technology,~\IEEEmembership{Staff,~IEEE,}
        % <-this % stops a space
% \thanks{This paper was produced by the IEEE Publication Technology Group. They are in Piscataway, NJ.}% <-this % stops a space
% \thanks{Manuscript received April 19, 2021; revised August 16, 2021.}

% The paper headers
% \markboth{Journal of \LaTeX\ Class Files,~Vol.~14, No.~8, August~2021}%
% {Shell \MakeLowercase{\textit{et al.}}: A Sample Article Using IEEEtran.cls for IEEE Journals}

% \IEEEpubid{0000--0000/00\$00.00~\copyright~2021 IEEE}
% Remember, if you use this you must call \IEEEpubidadjcol in the second
% column for its text to clear the IEEEpubid mark.
% \IEEEpubidadjcol
\maketitle

\begin{abstract}
Recently, due to the poor performance of supervised person re-identification (ReID) to an unseen domain, Domain Generalization (DG) person ReID has attracted a lot of attention which aims to learn a domain-insensitive model and can resist the influence of domain bias. In this paper, we first verify through an experiment that style factors are a vital part of domain bias. Base on this conclusion, we propose a Style Variable and Irrelevant Learning (SVIL) method to eliminate the effect of style factors on the model. Specifically, we design a Style Jitter Module (SJM) in SVIL. The SJM module can enrich the style diversity of the specific source domain and reduce the style differences of various source domains. This leads to the model focusing on identity-relevant information and being insensitive to the style changes. Besides, we organically combine the SJM module with a meta-learning algorithm, maximizing the benefits and further improving the generalization ability of the model. Note that our SJM module is plug-and-play and inference cost-free. Extensive experiments confirm the effectiveness of our SVIL and our method outperforms the state-of-the-art methods on DG-ReID benchmarks by a large margin.
\end{abstract}


\section{Introduction}
\label{sec:intro}
% 
% \lorem{1}
% \input{fig/teaser}
% \lorem{2}

\begin{figure}
\begin{center}
% 这个图的构图后续还需要改，但是主要内容不会变了
\includegraphics[width=1.\columnwidth]{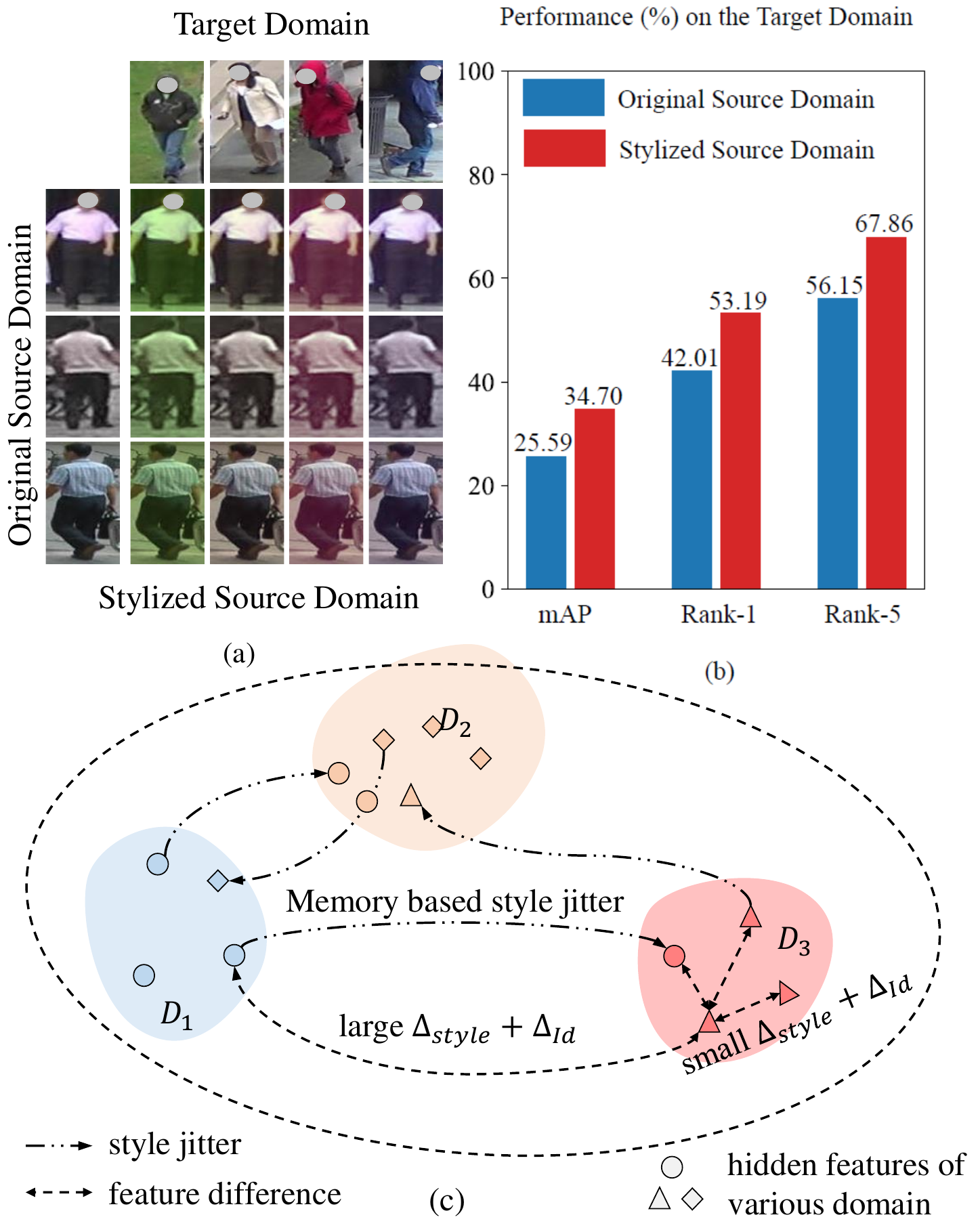}
\end{center}
% \vspace{-6mm}
\caption{
Illustration of experiments exploring the influence of style factors in domains bias.
(a) The exploratory experiment introduces the style information of the target domain into the original source domain and then obtain the stylized source domain.
(b) The experiments results of the model trained on original/stylized source domain and tested on the target domain.
The performance is obviously improved after introducing style information.
% , meaning that style factors play an important role in DG-ReID.
(c) The visualization of hidden features affected by style jitter module.
% \textcolor{red}{domain bias of DG-ReID.}
}
\label{fig:intro}
\end{figure}

% Person re-identification (ReID) aims to identify a specific person across non-overlapping cameras under various locations and views.
Person re-identification (ReID) aims to identify a specific person across non-overlapping cameras under various views and scenes. 
It has attracted a lot of attention due to its significance to intelligent surveillance systems. 
With the development of deep learning techniques, person ReID has achieved remarkable performance in supervised manner, where a model is trained and tested on the same dataset \cite{sun2020circle, Chu_2019_ICCV, he2021transreid, wu2016personnet, qian2017multiscale, liu2018ram, wang2018cosface}. 
% 这里对 domain 和 dataset 的关系论述顺序存疑；后面需要看
Unfortunately, due to the dramatic domain bias, these ReID models suffer from severe performance degradation when applying to an unseen domain. 
To cope with this issue, tasks of unsupervised domain adaption (UDA) and domain generalization ReID (DG-ReID ) have attracted lots of attention in recent years.
In UDA task, the images of target domains are available but these labels are unknown.
These approaches \cite{ge2020mutual, song2020unsupervised, fu2019self, kumar2020unsupervised} generally adapt the source-trained model and utilize the unlabeled target data to finetune.
% and adapt the source-trained model to the target domain.
Whereas in DG-ReID, both images and labels are invisible, which makes the task more challenging but practical.
% the images of target domain are available, and the labels are unavailable.
% Therefore, unsupervised domain adaption (UDA) methods have been studied, where unlabeled target domain data can be utilized to adapt the model from source to target domain [XXX]. 
% But sometimes target domain's data is hard to collect and need to inference directly. 
%But the UDA methods need target domain data to fine-tune the model. Sometimes target domain's data is hard to collect and fine-tuning the model is also time-consuming. 
% To alleviate the above issues, the task of domain generalization (DG) has drawn growing attention from researchers. Different from DG classification method which use united classifier to discriminate different categories, DG-ReID is more challenging due to the label space of different dataset are disjoint. 
% Existing DG-ReID 
% methods can be divided into three main aspects, adversarial learning based methods[XXX] employ  adversarial auto-encoder module and discrimination task to help extract domain-invariant features but usually imbalance between several tasks, Normalization-based methods[XXX] introduce instance normalization(IN) to disentangles identity-relevant and identity-irrelevant features but only focus on single sample. Meta-learning based methods[]   adopted the training paradigm to split the train datasets into meta-train and meta-test to simulate the domain bias, This has proved effective but model is easy to overfit the bias during later stage of training.

Existing DG-ReID methods can be roughly divided into three main aspects.
Adversarial learning based methods \cite{lin2020multi, tamura2019augmented} employ the adversarial auto-encoder module and discrimination task to help extract domain-invariant features, but these methods may fall into the unbalanced training between several tasks.
Normalization based methods \cite{jin2020style, choi2021meta, li2018domain} utilize the instance normalization (IN) to disentangle the identity-relevant and identity-irrelevant features.
They generally focus on a single sample and ignore the relationship between multiple samples, including intra-domain and inter-domain.
Meta-learning based methods \cite{finn2017model, li2018learning, balaji2018metareg} split the train datasets into meta-train and meta-test to simulate the domain bias, adopting a process of "learning-to-learn" to optimize the model. 
Although the goal of meta-learning is to prevent the model from overfitting, as the training process progresses, the model will witness enough data (including the data in the meta-test) and then inevitably fall into overfitting.
% the training paradigm to split the train datasets into meta-train and meta-test to simulate the domain bias, This has proved effective but model is easy to overfit the bias during later stage of training.
% In this paper, we consider a basic problem —— “what factors does domain bias contain?”, it may include scene, season, camera angle and lighting etc. How much is the influence of style factor(e.g., illumination, hue, contrast)? Therefore, we did a simple experiment as shown in Figure ?, we add the style of the target domain to the source domain in training phase by replace mean and variance, which [XXX] prove that the statistical value can be used as a representative of style. what shocked us was the generalization ability of the model has been significantly improved just through such a simple operation. it's true that in reality, we cannot predict the style of the target domain in advance, but expand various styles is feasible.
Most of methods mentioned above are inspired by approaches in conventional DG task, which is a classification task and different from DG-ReID task.
Specifically, researchers typically define various datasets as different domains.
The datasets (domains) in conventional DG task may contains real image dataset and painting picture dataset, where the domain bias is obvious.
% The source domain in conventional DG task may be a dataset consisting of real images, and the target domain may be a dataset
% The source and target domains (datasets) have the same label space in conventional DG task.
% For instance, the source domain can be 
% Generally, various datasets can be viewed as different domains, respectively
% these approaches assume that the source and target domains have the same label space, 
However, DG-ReID task involves only real image datasets.
Images in different datasets (domains) are captured from different cameras and regions.

In this paper, we consider a basic problem: “what factors does domain bias contain in DG-ReID?”
% Generally, various datasets can be viewed as different domains, respectively, where the images are captured from different cameras and regions.
% Therefore, the domain bias (\textit{i.e.}, the differences between various datasets) may include scene, season, camera angle and lighting etc. 
% Therefore, the domain bias (\textit{i.e.}, the differences between various datasets) may include scene, season, camera angle and style (\textit{e.g.} illumination and hue). 
% stylexx 具体来说，这些style包含
% \textcolor{red}{These factors will definitely cause a change in the style of the images, \textit{e.g.} illumination, hue and contrast.}
An intuitive idea is that the domain bias (\textit{i.e.}, the differences between various datasets) may include scene, season, camera angle and style (\textit{e.g.} illumination and hue). 
To explore the influence of style factors in domain bias, we conduct a simple exploratory experiment as illustrated in Fig. \ref{fig:intro}.
% 看情况加 fig1
Specifically, by replacing the mean and variance statistics of the images, in Fig. \ref{fig:intro} (a) we introduce the style information of the target domain (DukeMTMC \cite{ristani2016performance}) into the source domain (Market 1501 \cite{zheng2015scalable}), where the mean and variance are considered and proved as style representation in the transfer learning task \cite{huang2017arbitrary, liu2021adaattn, tang2021crossnorm}.
The results in Fig. \ref{fig:intro} (b) shows that we can achieve better ReID performance by training on the source domain with the same style as the target domain, confirming that the style factors are a vital part of domain bias.
% The experiment of introducing the style information of the target domain achieve a great performance improvement.
However, style information of the target domain is unavailable in DG-ReID.
% However, we can not obtain the style information of target domain in DG-ReID.
This makes us have to eliminate the influence of style factors to reduce the domain bias.

Inspired by the above observations, we propose a 
Style Variable and Irrelevant Learning (SVIL) method for DG-ReID.
We start our approach with multi-source DG-ReID task.
Different with multi-source DG classification task, the label spaces of various source domains in DG-ReID are different.
This means that an identity exists only in its specific domain, and identity and domain are strongly related.
As shown in Fig. \ref{fig:intro} (c), identity $x^{a}_{1}$ belongs to domain $D_{1}$, and identities $x^{b}_{3}$ and $x^{c}_{3}$ belongs to domain $D_{3}$.
Here we do not consider the influence of non-style domain bias and there are large style difference between cross-domain identities (\textit{e.g.,} $x^{a}_{1}$ and $x^{b}_{3}$).
Then models can easily distinguish these identities relying on the style bias, thereby ignoring the extraction of identity information and achieving poor generalization.
% 暂定，先画图
To prompt the model focusing on identity-relevant information and be insensitive to the style variations, our SVIL enhance the style diversity of the specific source domain and reducing the style differences of different source domains.
% By enhancing the style diversity of the specific source domain and reducing the style differences of different source domains, our SVIL can prompt the model to focus on identity-relevant information and be insensitive to the style variations.
% which is generalizable to unseen target domain.
% The model is style-insensitive and 
% enhance style-irrelevant and identification-relevant feature learning. 
% SJM 的low level 的
More concretely, we design a Style Jitter Module (SJM) to enrich the style diversity of source domains.
% The SJM module can \textcolor{red}{focus on style features in various scales}, changing style of features with the help of style bank.
% The SJM module can extract the style representations of different scale features, and fuse them into other features with the help of a style bank.
The SJM module uses style memory to store identity style statistics in all source domains.
Then for each identity in mini-batch, we generate a new style statistic based on all statistics, where the new style is a weighting of the negative sample styles in all domains based on the identity relationship modeling.
The original style is replace by the new style to destroy the relationship between original style and identity.
To better distinguish hard negative samples, we strengthen the importance of the styles of hard samples during style generation by hard identity emphasis.
Besides, we perform cross-domain identity emphasis strategy to avoid the information of identities within the same domain to dominate the generated styles.
% It also makes cross-domain negative samples become harder, which is conducive to the model's pay more attention to identity features.
Our SJM module is a inference cost-free module, without learnable parameters.
It leads to variable styles between training samples and makes negative samples harder.
% It also makes cross-domain negative samples become harder, which is conducive to the model's pay more attention to identity features.
% 补充 SJM 细节

In addition, we notice that previous methods \cite{jin2020style,zhao2021learning,dai2021generalizable} employ a unified loss functions on multiple domains to learning feature representations.
They omit the strong relationship between identity and domain caused by various label spaces.
To weaken the relationship, we propose domain-agnostic loss and domain-specific loss to jointly learn robust features.
% The two losses a
% We also apply tow kinds of loss functions to supervise the learning of network. design 
A meta collaborative training procedure is employed as well to maximize the effectiveness of our SJM module, where the meta-learning algorithm is organically combined.
% We combine a Model-Agnostic Meta Learning (MAML) algorithm with our SJM module to maximize its effectiveness. 
% Different from above mentioned meta learning methods, the meta collaborative training with 
The SJM module can enlarge and continuously change the domain bias between meta-train and meta-test, successfully delaying the model from overfitting.
% making learning process not easy to overfit.
% Different the 
% By simulate the domain bias between meta-train and meta-test datesets, MAML adopts the concept of "learning-to-learn" to produce a generalizable model.
% 与上面MAML的缺点形成 Callback
% Our SJM module can enlarge and continuously change the domain bias, making learning process harder.
The combination of these two sub-method produce a synergy effect, leading to a style-irrelevant model to generalize to the unseen target domain.
% is generalizable to unseen target domain.
% It needs to be explained that our SJM module is a plug-and-play module and inference cost free.
% \textcolor{red}{It should be explained that our SJM module is a plug-and-play and inference cost free module.}
% Under general settings, feature learning is limited to a few hard negative samples of the same source domain due to the style difference between domains is obviously, XXX can significantly reduce the style gap and eliminate the binding relationship between person identification and dataset, because the style of the same sample(image or feature) is constantly changing during training. And Meta-Learning can XXX. more detail, Style Random Jitter Module (SJM) was designed, SJM Randomly use the mean and variance of the other sample in the style bank to cover the original sample, which also can use multiple styles into different local part. It needs to be explained that SJM is plug-and-play and inference cost free. Through these efforts, we can achieve
% this effect that features of positive samples are more abundant and more negative samples contribute to training.

% DG aims to learn a generalizable model for unseen domains without having to access the target domain data.
% 

We also perform the proposed SVIL on the single-source DG-ReID task to confirm its effectiveness.
Specifically, because images captured under different cameras have stylistic differences, we divide a single dataset into multiple sub-datasets according to cameras.
Then we employ SVIL on the single-source DG-ReID task to learn style-insensitive model as we do on the multi-source DG-ReID task.
In summary, the major contributions of our works are summarized as follows:
\begin{itemize}
% [leftmargin=*]
% \setlength\itemsep{-.3em}
% 发现
\item Based on the observation that style factors play a vital role in DG-ReID, we propose a Style Variable and Irrelevant Learning (SVIL) method to eliminate the influence of style variations and ah.
% \item Based on the observation that style factor is important for domain generalization, We propose a ... method  % style 在 DG-ReID 上的辨析
% SJM模块
\item We propose a Style Jitter Module (SJM) to generate diverse styles for identities, increasing the variety of feature styles.
It forces the model to focus on identity-relevant information and ignore style-relevant information.
% \item We propose Style Random Jitter (SJM) module to enhance the richness of feature style, make the model to focus on identity-relevant information. more over, SJM module is a plug-and-play and inference cost free module. % SJM module
% \item We propose a meta-learning based [title/method-name] framework % SJM module 与 Meta 结合的方法
% 结合 Meta Learning 能达到 sota
\item We apply a meta collaborative training procedure to get rid of the influence of style further.
% and they produce a synergy effect.
Our approach outperforms state-of-the-art methods by a large margin in both multi-source and single-source DG-ReID tasks.
% \item We propose to combine SJM with meta-learning framework to enlarge the domain shift in order to avoid overfitting to source domain bias. Our method outperforms state-of-the-art DG ReID methods by a large margin.
% \item Combine meta-learning framework, we evaluate our method on various DG ReID benchmarks and our method outperforms state-of-the-art DG ReID methods by a large margin. % 实验
\end{itemize}

The rest of this paper is organized as follows. 
Section \ref{sec:related} reviews and discusses the related works.
Section \ref{sec:method} first introduce s the overall framework of the proposed method, and then elaborate the designed style jitter module and the training procedure.
Section \ref{sec:experiment} presents the experimental results.
Finally, Section \ref{sec:conclusion} draws the conclusion.
\section{Related works}
\label{sec:related}

% \AT{introduce subtopics, mention surveys here}
% \lorem{1}

\textbf{Re-identification}. 
Re-identification (ReID) tasks have attracted a lot of attention due to their wide application in urban surveillance and intelligent transportation \cite{ye2021deep, khan2019survey}. 
Both person ReID and vehicle ReID have achieved significant performance through supervised learning approaches. 
For instance, Sun \textit{et al.} \cite{sun2020circle} propose a circle loss function for deep feature learning. 
Chu \textit{et al.} \cite{Chu_2019_ICCV} propose a novel viewpoint-aware metric learning approach for vehicle Re-ID. 
He \textit{et al.} \cite{he2021transreid} design a pure transformer framework to ReID tasks and achieve remarkable performance.
Other various attempts \cite{wu2016personnet, qian2017multiscale, liu2018ram, wang2018cosface} along supervised learning acquire great performance as well. 
However, all of them pay little attention to exploring invariant information across domains, which limits their effects in practical applications.
Therefore, Unsupervised Domain Adaptation (UDA) methods \cite{ge2020mutual, song2020unsupervised, fu2019self, kumar2020unsupervised} are proposed to alleviate differences between the source domain and target domain. 
For instance, Ge \textit{et al.} \cite{ge2020mutual} propose a MMT framework to tackle the noise label in clustering process. 
Song \textit{et al.} \cite{song2020unsupervised} employ a self-training scheme to iteratively minimize the loss functions. 
These methods require enough unlabeled training data of target domain, which is unsatisfactory in many practical situations.
Then the domain generalization task is established.
% Recently, re-identification (ReID) tasks have attracted a lot of attention due to its wide application in urban surveillance and intelligent transportation \cite{ye2021deep, khan2019survey}. Both pedestrian ReID and vehicle ReID have achieved significant performance through supervised learning approaches. Sun \textit{et al.} \cite{sun2020circle} proposed circle loss function for deep feature learning. Chu \textit{et al.} \cite{Chu_2019_ICCV} proposed a novel viewpoint-aware metric learning approach for vehicle Re-ID. He \textit{et al.} \cite{he2021transreid} proposed a pure transformer framework to ReID tasks and achieved remarkable performance, and other various attempts \cite{wu2016personnet, qian2017multiscale, liu2018ram, wang2018cosface} along supervised learning have led to great performance. But all of them paid little attention on exploring invariant information across domains, which limits their effects in practical applications. Therefore, unsupervised domain adaptation (UDA) methods are proposed to alleviate differences between source domain and target domain. Ge \textit{et al.} \cite{ge2020mutual} proposed MMT framework to tackle noise label in clustering process. Song \textit{et al.} \cite{song2020unsupervised} proposed a self-training scheme to iteratively minimize the loss functions. But these and other methods \cite{fu2019self, kumar2020unsupervised} need to get enough unlabeled training data from target domain, which cannot be met in many practical cases.

\textbf{Domain Generalization.} 
The goal of Domain Generalization (DG) is to learn a model that is generalizable to unseen domains, having received more and more attention from the research community \cite{muandet2013domain, tang2021crossnorm, zhou2021domain}. 
% Compared to unsupervised learning and unsupervised domain adaptation, DG is a more challenging problem where the target domain data is unavailable. 
% Due to close to reality, domain generalization \cite{muandet2013domain, tang2021crossnorm, zhou2021domain} has received more and more attention from the research community. 
% The above DG 
These methods generally assume that the source and target domains have the same label space. 
Whereas in DG-ReID, the label spaces are non-overlap.
This results in methods in conventional DG can not be applied directly to DG-ReID.
% ReID is an open-set problem where label spaces between different domains are disjoint, thus these general DG methods can not be applied directly.
% Different from the conventional DG
% \indent In terms of Re-ID DG, the adversarial learning based method is a popular method. Lin \textit{et al.} \cite{lin2020multi} proposed a multi-dataset feature generalization network, which is capable of learning a universal domain-invariant feature representation from multiple labeled datasets and generalizing it to unseen camera systems. Tamura \textit{et al.} \cite{tamura2019augmented} proposed augmented hard example mining, which can be easily integrated to a common Re-ID training process and can utilize sophisticated models without any network modification. However, the performance of these methods varies greatly in different tasks, and it is difficult to obtain a stable model. 
Adversarial learning is a popular method \cite{lin2020multi, tamura2019augmented} used in DG-ReID. 
For instance, Lin \textit{et al.} \cite{lin2020multi} propose a multi-dataset feature generalization network to learning a universal domain-invariant feature representation and generalize it to unseen camera systems. 
% Tamura \textit{et al.} \cite{tamura2019augmented} propose the augmented hard example mining, which can be easily integrated to a common Re-ID training process. However, their method is not stable on different tasks.
Normalization-based method is also a mainstream solution in DG-ReID. 
For instance, Jin \textit{et al.} \cite{jin2020style} design a Style Normalization and Restitution (SNR) module to enhance generalization capabilities of networks. 
Choi \textit{et al.} \cite{choi2021meta} propose the MetaBIN framework which prevents models from overfitting to the given source styles and improves the generalization capability to unseen domains.  
% Li \textit{et al.} \cite{li2018domain} proposed to align the distributions between different domains using Maximum Mean Discrepancy (MMD) measure. 
Compared to these methods, our SVIL can focus on more detailed style information.
% These approaches also have some shortcomings that due to much attention paid to the features of the instance normalized samples, the model may lack of the ability to find subtle features of samples. 
Beside, Model-Agnostic Meta-Learning (MAML) \cite{finn2017model} is another popular method for DG-ReID. 
Zhao \textit{et al.} \cite{zhao2021learning} propose a memory-based framework to solve DG-ReID, where the meta-learning algorithm is adopted.
Similarly, Dai \textit{et al.} \cite{dai2021generalizable} also employ a vote experts under the meta-learning framework.
% MAML \cite{finn2017model} was initially proposed to find a good initialization of parameters for novel tasks in conventional DG.
% Recently, MAML-based methods have also been successfully applied in DG area. 
% Li \textit{et al.} \cite{li2018learning} proposed MLDG, a MAML-based method for DG. 
% Balaji \textit{et al.} \cite{balaji2018metareg} proposed to encode the notion of domain generalization using meta-learning framework. 
% However, in most cases, meta-learning always suffer from overfitting to multiple training sets.
In our method, the SJM module can contribute to the strengths of meta-learning, produce a synergy effect.

\textbf{Style Transfer.}
In the field of style transfer, it has been proved that the statistical information of the features extracted by DNN can represent the style of the image \cite{gatys2016image, li2016combining, li2017demystifying}. 
Gatys \textit{et al.} \cite{gatys2016image} use second-order statistical features as the optimization goal to achieve style transfer. 
Li \textit{et al.} \cite{li2017demystifying} find that other statistics (\textit{i.e.,} mean and variance) of Batch Normalization (BN) layers contain the traits of different domains. 
Ulyanov \textit{et al.} \cite{ulyanov2017improved} utilize Instance Normalization (IN) to obtain the statistical characteristics of features. 
Dumoulin \textit{et al.} \cite{dumoulin2016learned} proposed Conditional Instance Normalization (CIN), which would learn pairs of $\alpha$ and $\beta$ parameters during training. 
When both image content and decoding network are the same, adopting different pairs of $\alpha$ and $\beta$ will get different styles of transfer results. 
In IN and CIN, the network can learn the affine transformation parameters $\alpha$ and $\beta$.
In addition, to remove the two learnable parameters, AdaIN \cite{huang2017arbitrary} are proposed.
It directly replaces the two parameters with the mean and standard deviation of the style image features.

\section{Method}
\label{sec:method}
% The purpose of our approach is to enhance the generalization ability of the model to directly.

% This purpose of our approach is to improve the style insensitivity of the model to increase its generalization ability.

% Improve the style insensitivity of the model to increase the generalization of the model

% This work aims to introduce a generalizable framework on 
% In this work, we aim to learn a generalizable model which can be deployed directly to a new unseen domain. 
% By increasing the domain diversity during the training, the network will be insensitive to domain variations.
% This lead to stronger generalization ability of model on new domain.
% This work aims to learn a generalizable model that can be deployed directly to a new unseen domain. 
% This purpose of our method is to address the style variations problem in DG ReID to narrow the domain gap.
This purpose of our method is to narrow the domain gap caused by style variations in different domains.
By increasing the style diversity on the training stage, the network will be insensitive to style changes between different domains and pay more attention to identity-relevant information.
% Then the domain gap is shrunken a bit
% The identity-relevant information is emphasized
% So 
In this way, we can obtain a more generalizable model that can be deployed directly to a new unseen domain.
% it can lead to a more vital generalization ability of the model on the new domain.
% ...

% \begin{figure*}
% \begin{center}
% \includegraphics[width=.99\columnwidth]{example-image-golden}
% \hfill
% \includegraphics[width=.99\columnwidth]{example-image-golden}
% \end{center}
% \caption{
% % 
% \textbf{Outline -- }
% A figure summarizing the entire algorithm
% }
% \label{fig:pipeline}
% \end{figure*}
\begin{figure*}
\begin{center}
% 后续再换 pdf 格式, 需要需要进一步修改
\includegraphics[width=2\columnwidth]{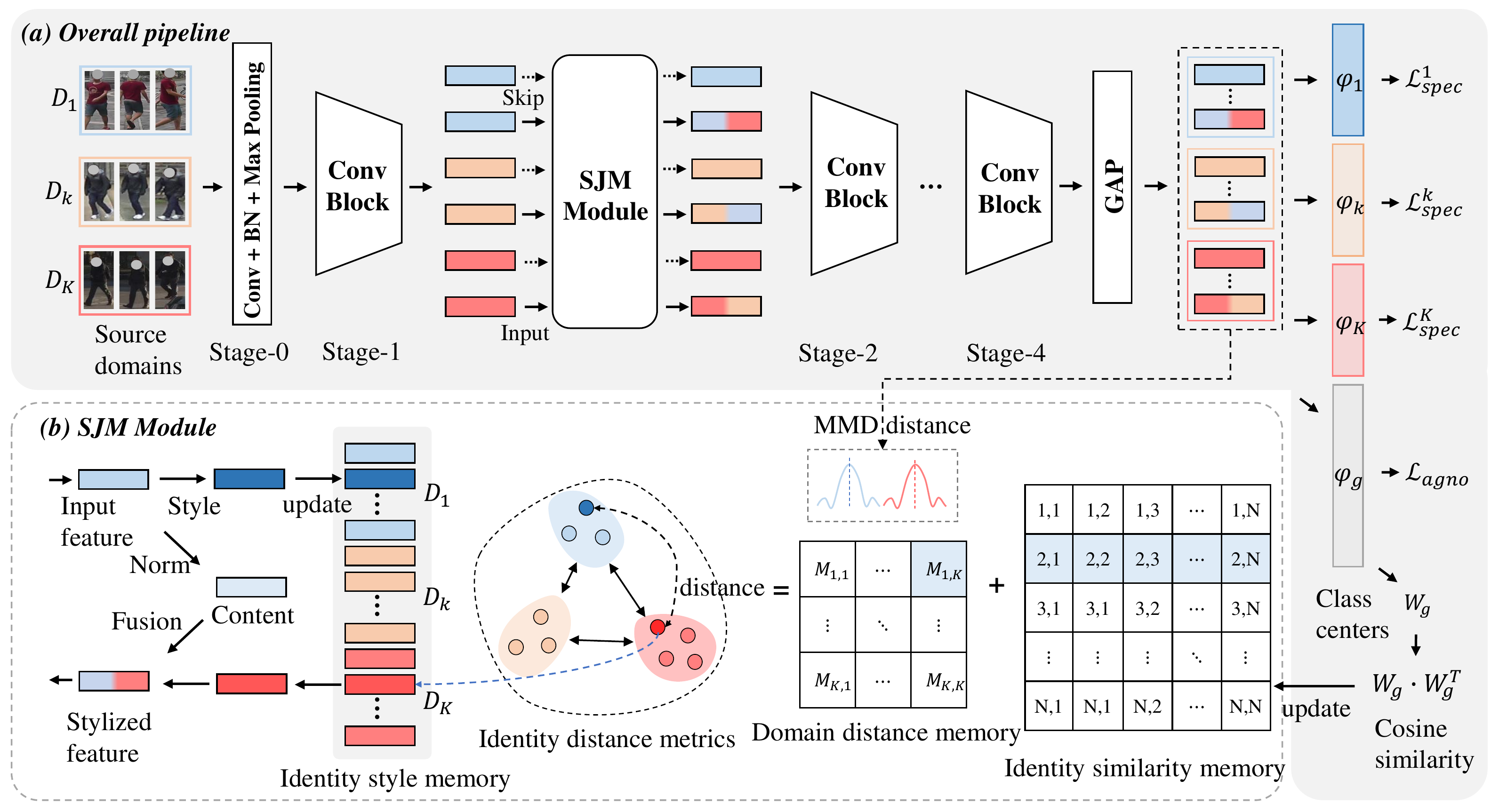}
\end{center}
\caption{
Illustration of our method.
(a) The overall pipeline of our method.
The proposed SJM module can be plugged after any stage of the backbone (\textit{e.g.}, ResNet50), and we illustrate the structure of SJM between stage-1 and stage-2.
In the training stage, we sample images from $K$ source domains and group them into a mini-batch.
Then images are fed into the backbone network and pass through the SRJ module, in which the style of features are changed within and across domains with the help of a style bank.
After the Global Average Pooling (GAP), stylized features are utilized to calculate losses.
% The stylized features are enforced different domain-specific loss $\mathcal{L}_{spec}$ according to their domain.
Features from the same domain in the mini-batch are applied the same domain-specific loss $\mathcal{L}_{spec}$.
All of the features calculate the domain-agnostic loss $\mathcal{L}_{agno}$.
% Our Style Random Jitter (SRJ) module can be plugged after any stage (\textit{e.g.}, stage-1) of the backbone (\textit{e.g.}, ResNet50) in the training stage, where the style of features are changed within and across domains.
% These features pass through the backbone
}
\label{fig:pipeline}
\end{figure*}
% \vspace{-3mm} %< bloody latex and its heuristics for figure placement
\subsection{Overview}
\label{sec:overview}
% In this work, we aim to learn a generalizable model which can be deployed directly to a new unseen domain. 
% To this end, we propose a [method]. 
% \textcolor{red}{[tag]}  
The main pipeline of our method is illustrated in Fig. \ref{fig:pipeline} (a). 
In the train stage, we can access $K$ source domains $\mathcal{D}=\{\mathcal{D}_k\}_{k=1}^K$, where $\mathcal{D}_k=\{({x}_i^k,y_i^k)\}_{i=1}^{N_k}$ and $({x}_i^k,y_i^k)$ is a image-label pairs belong to $\mathcal{D}_k$, 
$N_k$ is the number of images in the source domain $\mathcal{D}_k$.
For each sample $x_i^k$, its label $y_i^k$ comes from the specific label space 
$\mathcal{Y}_k$.
% $\mathcal{Y}_k = \{1,2,\dots,M_k\}$.
% 待续：补充一句对 dgr-reid label space 性质的说明，然后再引出下文的全局 label
By combining all source domains (termed global domain), we can obtain its global label ${y}_i^g \in \mathcal{Y}_g$, where
$\mathcal{Y}_g$ is the global label space.
% $\mathcal{Y}_g = \{1,2,\dots,M_g\}$ and $M_g= \sum_{k=1}^{k}{M_k}$.
% Images from different source domains form a mini-batch 
% The batch size is $B \cdot K$, indicating that $B$ image $x_i^k$ from each domain $\mathcal{D}_k$
We sample $B$ images $x_i^k$ from each domain $\mathcal{D}_k$, where $B=\mathcal{P} \cdot \mathcal{K}$, meaning $\mathcal{K}$ images from each of $\mathcal{P}$ person identities.
Then all images form a mini-batch of size $B \cdot K$ for all domains ($\mathcal{P} \cdot K$ identities in total).
They are fed into the backbone network $f(\cdot)$ (\textit{e.g.,} ResNet-50).
After the Global Average Pooling (GAP), we design $K$ domain-specific classifiers $\varphi_{k}(\cdot)$ and a domain agnostic classifier $\varphi_{g}(\cdot)$ to learn the feature representations.
% we design a memory-based hybrid classifier $\varphi(\cdot)$, where the losses are calculated.
% 可能不使用 memory
% where the domain specific loss $\mathcal{L}_{\text{spec}}$ and the domain agnostic loss $\mathcal{L}_{\text{agno}}$ are calculated.
% 
% As shown in Fig. \ref{fig:pipeline}, 

We plugged the SJM module between stage-1 and stage-2.
During the training phase, one half of the images of each identity pass through the SJM module, and the style information of feature representations in stage-1 is changed.
The style of the other half of the images remain as it as.
where the style information of different feature representations is changed. 
Next, all of the features are fed into stage-2 until the end of the network.
We enforce domain-specific loss $\mathcal{L}_{\text{spec}}$ on each specific domains' features and predictions. 
To make the model more domain-insensitive, a domain-agnostic loss $\mathcal{L}_{\text{agno}}$ is employed on the global domains' features and predictions as well.
Two losses jointly prompt the network to ignore the style variation of features and focus on identity-relevant information.
In addition, we apply a meta-learning algorithm combined with the proposed SJM module to training the network, which can produce a synergy effect to our method (more details in Sec. \ref{sec:Optimization}).
Our SJM module is only used for training and will be discarded in testing.

\subsection{Style Jitter Module}
\label{sec:SJM}

Images of different domains in DG ReID are generally captured from different cameras and scenes, leading to various styles.
% To address the style bias in domain bias, 
To address the style variations in DG ReID,
we propose a SJM module to decreases the model's sensitivity to style variations.
% The SJM module is a plug-and-play and inference cost-free module, without learnable parameters.
The SJM module takes one hidden representation from a specific domain as the input and generate a stylized representation, in which the cross-domain style information is transferred.

% % 在这句之前，必须把 dg reid 是一个 heterogeneous 问题解释清，为后面铺垫
% % 1. 能在 local 范畴内，让模型聚焦于本身，而非style;
% % For one thing, the SJM module can introduce style information of other domains to a specific domain, increasing its style diversity.
% % 另一方面，SJM能为不同domain创造共同信息，在global domain范畴内使model聚焦于 feature 本身，而不是 style 信息；
% % For another, it 
% On the one hand, the SJM module can enrich the style diversity of each identity, improving the model's tolerance for style changes.
% On the other hand, it is able to introduce styles from other domains into specific domains, increasing the number of hard samples.
% Both contributes can brings a style-insensitive model that is absorbed in identity features and performs better in an unseen target domain.
% % generalized model, which is insen
% % By reducing the style bias of different domains, our model are more generalize to unseen target domain.

Specifically, we denotes backbone network as $f(x)=f_{m}(g_{m}(x))$, where $g_{m}$ denotes the part of the network mapping the input data $x$ to the hidden representation $F = g_{m}(x) \in \mathbb{R}^{H \times W \times C}$ after the $m$-th stage (\textit{e.g.}, $m=1$ in Fig \ref{fig:pipeline}) and $f_{m}$ denotes the part of the network mapping the feature $g_{m}(x)$ to the feature vector after the GAP layer.
In a mini-batch including $\mathcal{P} \cdot K$ identities and each identity including $\mathcal{K}$ images, we can obtain their maps at $m$-th hidden stage.
For each identity, we select $\frac{\mathcal{K}}{2}$ representations $F$ and diverse them by SJM module.
As shown in Fig. \ref{fig:pipeline} (b), we first compute the channel-wise mean and standard deviation $\mu(F), \sigma(F) \in \mathbb{R}^C$ by the style operation as follows:
\begin{normalsize}
\begin{align}
\mu({F}) &= \frac 1 {HW} \sum_{h=1}^H \sum_{w=1}^W {F}_{hw},
\end{align}
\end{normalsize}
\begin{normalsize}
\begin{align}
\sigma({F}) &= \sqrt{ \frac 1 {HW} \sum_{h=1}^H \sum_{w=1}^W ({F}_{hw} - \mu({F}))^2 + \epsilon},
\end{align}
\end{normalsize}
These feature statistics can capture informative characteristics of the specific domain and be viewed as style representation according to previous works \cite{huang2017arbitrary}.
Then we construct a identity style memory to store the style representations of each identity, where the cross-domain style representations are generated.

\subsubsection{Identity Style Memory Construction}
% \textbf{Identity Style Memory.} 
The style memory maintain the style representation of all identities. 
For $|\mathcal{Y}_g|$ identities from all source domains, the memory $\mathcal{\bar M}$ has $|\mathcal{Y}_g|$ slots, where each slot save the style features of the corresponding identity.
In initialization, we utilize the model to extract style representations for all source samples.
Then we initialize the representation $\mathcal{\bar M}[j]=(\bar \mu[j], \bar \sigma[j])$ in memory using the average of all representations for $j$-th identity.
At each training iteration, we update the memory with the style representations in the current mini-batch, which is formulated as:
\begin{normalsize}
\begin{equation}
\mathcal{\bar{\mu}}[j] \leftarrow m \cdot \bar{\mu}[j]+(1-m) \cdot \frac{2}{|\mathcal{B}_j|} \sum_{x_i \in \mathcal{B}_j} \mu (g_{m}(x_{i})),
\end{equation}
\end{normalsize}
\begin{normalsize}
\begin{equation}
\mathcal{\bar\sigma}[j] \leftarrow m \cdot {\bar\sigma}[j]+(1-m) \cdot \frac{2}{|\mathcal{B}_j|} \sum_{x_i \in \mathcal{B}_j} \sigma (g_{m}(x_{i})),
\end{equation}
\end{normalsize}
where $g_{m}(x_{i})$ is the input feature $F$, $\mathcal{B}_j$ refers to the samples belonging to the $j$-th identity and $|\mathcal{B}_j| = \mathcal{K}$ is the number of samples for the $j$-th identity in the current mini-batch.
$m \in [0,1]$ determines the update rate.

% \subsubsection{Identity Relationship Modelling}
\subsubsection{Stylized Feature Generation} 
To introduce information from other domains into the current specific domain,
we generate new style representations by weighting all identity styles except the identities to which the input features belong.
More concretely, we obtain a identity-related weight $\alpha \in \mathbb{R}^{|\mathcal{Y}_g|}$ for the input feature by identity relationship modelling (more detail in Sec. \ref{sec:weight}).
Then the new style representation is achieved by weighted fusion of the features of all identities as follows:
\begin{normalsize}
\begin{equation}
\mu'= \sum _{j=1}^{|\mathcal{Y}_g|} \alpha_j \bar\mu[j],
\end{equation}
\end{normalsize}
\begin{normalsize}
\begin{equation}
\sigma'= \sum _{j=1}^{|\mathcal{Y}_g|} \alpha_j \bar\sigma[j],
\end{equation}
\end{normalsize}
where $\bar\mu$ and $\bar\sigma$ are style representations in memory.
Then we replace the style representation with the synthetic one to achieve the domain-agnostic feature, with is formulated as:
\begin{normalsize}
\begin{equation}
F' = \underbrace{\sum _{j=1}^{|\mathcal{Y}_g|} \alpha_j \bar\sigma[j]}_{\sigma'} \left(\frac {F - \mu(F)} {\sigma(F)}\right) + \underbrace{\sum _{j=1}^{|\mathcal{Y}_g|} \alpha_j \bar\mu[j]}_{\mu'}.
\end{equation}
\end{normalsize}
%

% In addition, the feature map $F$ is normalized by the channel-wise mean-variance normalization ($\operatorname{Norm}$), which is formula as follows:
% %
% \begin{align}
% \hat{F} = \operatorname{Norm}(F) =  \frac {F - \mu(F)} {\sigma(F)}, \label{eq:Norm}
% \end{align}
% %

\subsection{Identity Relationship Modeling}
\label{sec:weight}

The identity-related weights control synthetic style representations and influence stylized features.
Intuitively, for the features of a specific domain $\mathcal{D}_k$, on the one hand, we expect the generated features to contain more style information of other domains $\mathcal{D}_{i,i \ne k}$, which is conducive to learning domain-independent information. 
On the other hand, the synthesized features should be as difficult as possible, which is beneficial to improve the generalization ability of the model.
% To this end, we 
% 直观上，对于domain1的特征，一方面我们期望生成的特征包含更多其他domain的风格信息，这有利于学习domain无关信息。另一方面，我们希望合成的特征尽可能得困难，这有利于提高模型的泛化能力。
% \\
% \hspace*{\fill} 
% \\
\subsubsection{Hard Identity Emphasis}

% \textbf{Hard Identity Emphasis.}
We obtain difficult synthetic features by emphasizing the stylistic features of hard identities, \textit{i.e.,} larger identity-related weight $\alpha$.
As shown in Fig. \ref{fig:pipeline} (a), we employ a domain-agnostic classifier $\varphi_{g}(\cdot)$ to classify all features in a mini-batch, where the domain-agnostic loss is adopted as follows:
\begin{equation}
\normalsize
    \mathcal{L}_{agno} = \frac{1}{K}\sum_{k=1}^{K}\frac{1}{B}\sum_{i=1}^{B} \mathcal{L}_{ce}(\varphi_{g}(f(x_i^k)),{y}_i^g)+\mathcal{L}_{tri}(f(x),{y}^g),
    \label{eq:agno}
\end{equation}
where $\mathcal{L}_{ce}$ denotes the cross-entropy loss and $\mathcal{L}_{tri}$ denotes the triplet loss \cite{hermans2017defense}. 
We normalize the weight $W$ in $\varphi_{g}(\cdot)$ and the feature vectors $f(x)$, then the cross-entropy loss becomes as follow:
\begin{equation}
\normalsize
 \mathcal{L}_{ce} =
 -\log \frac{\exp( \cos(W_{y^g_i}, f(x_i^k)) /\tau)}{\sum_{j=1}^{|\mathcal{Y}_g|}{\exp( \cos(W_{j}, f(x_i^k)) /\tau)}},
 \label{eq:Softmax}
\end{equation}
in which $\cos(W_{j}, f(x)) = \frac{W^T_{j} \cdot f(x)}{||W^T_{j}|| \cdot ||f(x)||}$ is the cosine similarity between $W_{j}$ and $f(x)$,
% ${||W_{y^g_j}||}_2 = 1$ and ${||f(x_i^k)||}_2 = 1$, 
$\tau$ is a temperature learnable parameter.
Previous study \cite{NormSoftmax} shows that the normalized weight $W_{j}$ can be viewed as the center of features belonging to identity $j$.
Inspired by this, we use the similarity between identity centers to represent the similarity between identities, which is calculated as follows:
\begin{equation}
S = \frac{W \cdot W^T}{||W|| \cdot ||W||}
\label{eq:similarity}
\end{equation}
where $S \in \mathbb{R}^{|\mathcal{Y}_g| \times |\mathcal{Y}_g|}$ and $S_{i,j}$ denotes the cosine similarity between identities $i$ and $j$.
Then we can calculate the identity-related weight based on identity similarities.

\textbf{Identity Similarity Memory.}
At each training iteration, we cannot directly obtain the identity similarities of the current iteration in the SJM module.
Therefore we use the historical cumulative similarity to calculate the weight, which brings the added benefit that cumulative similarity is more representative of true identity similarity than current similarity.
More concretely, we build a identity similarity memory ${\bar S}$ to store the historical similarity via momentum updates.
% , which is similar to identity style memory.
At initialization, we set ${\bar S[j]} = (\frac{1}{|\mathcal{Y}_g|}, \dots, \frac{1}{|\mathcal{Y}_g|})$ to treat all identities equally and ${\bar S[j][j]}=0$ to avoid the influence of identity itself.

% We also build a identity similarity memory $\bar S$ to store the values.
% In initialization, the
% Similar to identity style memory, memory 

% \textbf{Soft weight Weighted.} 
Then for each input feature whose identity is $j$, we can obtain the soft identity-related weight based on the similarity memory as follows:
\begin{equation}
\alpha = softmax(\beta),
\label{eq:soft}
\end{equation}
where $\beta=\bar S[j]$ is the identity-related factor. The identity with high similarity to identity $j$ (\textit{i.e.}, the hard identity) has greater value, and its style representation are paid more attention.
We can also focus only on the hardest identity by using the hard weight:
\begin{equation}
\normalsize
\alpha_i = \begin{cases}
1, &i=\underset{k}{\operatorname{arg \, max}}(\beta)  \\
0, &i \ne\underset{k}{\operatorname{arg \, max}}(\beta)  
\end{cases},
\label{eq:hard}
\end{equation}
where the hard weight $\alpha \in \mathbb{R}^{|\mathcal{Y}_g|}$ is a one-hot vector.

According to the weight we can get the generated stylized features as described in Sec \ref{sec:SJM}.
All features in mini-batch are utilized to calculate the losses and update the network via gradient descent, including the weights $W$ in $\varphi_{g}(\cdot)$.
Then for each identity $j$, we set ${S}_{j,j} = 0$ and update the identity similarity memory:
\begin{equation}
\normalsize
\bar S[j] \leftarrow m \cdot {\bar S}[j]+(1-m) \cdot  {S}_j,
\end{equation}
in which ${S}$ is calculated by the last updated weight $W$.

\subsubsection{Cross-domain Identity Emphasis}
Although identity emphasis can make difficult identities get more attention, most of the difficult identities exist in the same domain, hindering the transfer of cross-domain information.
To emphasis the cross-domain identities, we consider the influence of different domains when calculating identity-related weights, \textit{i.e.,} give the cross-domain identities greater weights.

% More specifically, we further introduce the relation of the identity domains to calculate the identity-related weight.
More specifically, we further introduce the relation (distances) between different domains when computing the identity-related factor $\beta$.
As shown in Fig. \ref{fig:pipeline} (a), a feature set $A_k = {\{ f(x^k_i) \}}^B_{i=1}$ can be obtained for each source domain $\mathcal{D}_k$. 
Because the feature set $A_k$ contains only a few features and is a subset of the domain $\mathcal{D}_k$, its distribution cannot represent the distribution of the entire domain.
We model the distance between domains using the cumulative distribution distance between feature sets.
% We use the cumulative distribution distance (Maximum Mean Discrepancy (MMD) \cite{2006MMD}) between feature sets instead of the distance between domains.
Similar to identity similarity saving, we construct a domain distance memory $\bar D$ to store the cumulative MMD distances between various feature sets.
At each iteration, the distance between two sets is computed as follows:
% which is formulated as follows:
% 利用 MMD 计算domain 间距离
% standard MMD letex
% \begin{equation}
% \normalsize
% \operatorname{MMD}\left(X_{S}, X_{T}\right)=
% \left\|\frac{1}{\left|X_{S}\right|} \sum_{x_{s} \in X_{S}} \phi\left(x_{s}\right)-\frac{1}{\left|X_{T}\right|} \sum_{x_{t} \in X_{T}} \phi\left(x_{t}\right)\right\|
% \end{equation}
\begin{equation}
\begin{aligned}
\normalsize
D_{s,t} & = 
\operatorname{MMD}^2\left(A_{s}, A_{t}\right) \\
& =
{\left\|\frac{1}{\left|A_{s}\right|} \sum_{i=1}^B \phi\left(f(x^s_i) \right)-\frac{1}{\left|A_{t}\right|} \sum_{i=1}^B \phi\left(f(x^t_i) \right)\right\|}^2,
\end{aligned}
\label{eq:mmd}
\end{equation}
where $\phi(\cdot)$ is a particular representation that maps the feature $f(x)$ into a reproducing kernel Hilbert space. \ref{eq:mmd}
The memory $\bar D$ for $k$-th domain is updated as follows:
\begin{equation}
\normalsize
\bar D[k] \leftarrow m \cdot {\bar D}[k]+(1-m) \cdot  {D}_k.
\end{equation}

Given the identity $j$ of domain $\mathcal{D}_k$, we denote it as $\mathcal{D}^{-1}(j) = k$.
To enhance the cross-domain identities, the identity-related factor $\beta$ in Eq. \ref{eq:soft} and Eq. \ref{eq:hard} are redefined as follows:
\begin{equation}
\normalsize
\beta_i = \bar S[j][i] + \bar D[\mathcal{D}^{-1}(j)][\mathcal{D}^{-1}(i)],
\end{equation}
where $\beta \in \mathbb{R}^{|\mathcal{Y}_g|}$.
In this way, information about difficult identities and information about cross-domain identities are simultaneously emphasized.

% We use the cumulative MMD distance between feature sets instead of the distance between domains, where a domain distance memory is adopted.

\subsection{Training Procedure}
\label{sec:Optimization}

To make the model focus on domain-independent identity information, we adopt two losses with different effect scopes to supervise the learning of the network.
Furthermore, a Model-Agnostic Meta Learning (MAML) algorithm is applied to the training to maximize the effect of the proposed SJM module.
% In addition, a Model-Agnostic Meta Learning (MAML) algorithm is applied to 

\subsubsection{Loss Functions}

\textbf{Domain-agnostic Loss.}
% 待补充
Following the works \cite{zhao2021learning} in conventional multi-source DG task, we employ a domain-agnostic loss (Eq. \ref{eq:agno}) whose effect scopes are all source domains, \textit{i.e.,} global effect.
More concretely, 
% 全局影响指 给定一个anchor 样本 x，其负样本可能来自于任何一个domain。通过拉远负样本对间的距离，模型能学到他们的不同信息。然而，因为DG person reid 中存在的 身份-域 高相关性，在区分不同域的特征时，模型会更多的关注域信息，从而忽略了身份相关的域无关信息，降低了模型的泛化能力。
% 我们的SJM模型通过将 style 进行迁移，从而使模型关注到 身份相关信息；此外，我们也用了 domain-specific loss 来较少模型对 domain-relevant 信息的学习；
%
%
% 这一块还是需要修改的
global influence refers to given an anchor sample $x$, its negative samples may come from any domain.
By increasing the distance between pairs of negative samples, the model can learn the differences between them.
However, because the identity and domain in DG person reid are highly correlated, the model can easily distinguish the features of different domains through focusing on domain bias.
This makes the model ignore identity-related domain-independent information and reduces its generalization ability.
Our SJM module exchanges style information across domains, allowing the model to focus on identity-related information. 
In addition, we use a domain-specific loss to reduce the model's attention to domain-related information.
\\
\hspace*{\fill} 
\\
\textbf{Domain-specific Loss.}
Different from domain-agnostic loss, the effect scopes of domain-specific loss is a specific domain.
The specific loss for domain $\mathcal{D}_k$ is formulated as follows:
\begin{equation}
\begin{aligned}
\normalsize
\mathcal{L}^k_{spec} = \frac{1}{B}\sum_{i=1}^B \mathcal{L}_{ce}(\varphi_{k}(f(x_i^k)),{y}_i^k)+\mathcal{L}_{tri}(f(x^k),{y}^k).
\end{aligned}
\label{eq:spec-k}
\end{equation}
Then the overall specific loss is as follows:
\begin{equation}
\begin{aligned}
\normalsize
\mathcal{L}_{spec} = 
\frac{1}{K}\sum_{k=1}^K \mathcal{L}_{spec}^{k}.
\end{aligned}
\label{eq:spec}
\end{equation}
It means that the anchor sample and its negative sample must belong to the same domain, \textit{i.e.,} local effect.
Then model are not affected by domain bias and focus on the domain-irrelevant information.
In addition, SJM brings more styles to the features in the same domain.
More hard samples are generated, and the model becomes more discriminative.
Finally, the overall training loss is:
\begin{equation}
\normalsize
    \mathcal{L}_{all} = \lambda \mathcal{L}_{agno} + (1 - \lambda) \mathcal{L}_{spec},
    \label{eq:all}
\end{equation}

\subsubsection{Meta Optimizing}

% We have access to N source domain datasets. At each episodic training iteration, we randomly split the N datasets into M meta-train datasets and N-M meta-test datasets. For each domain in meta train datasets, we utilize SJM module to transform its style to other domains while still keep its semantic information unchanged. By changing the style of source training data while maintain its id label, we hope that the model relies more heavily on the more general semantic feature instead of domain relevant style feature when discriminates between pedestrians in different domains.\\
% \indent In our framework, we propose a meta-learning algorithm that trains the model to be more robust to domain shift. Specifically, we split the source domain datasets into meta-train and meta-test sets at each episodic training iteration. We vary the style of images in meta-train sets by SJM module and keep the style of images in meta-test sets unchanged. This is to simulate real train-test domain shifts so that we can train a model that can generalize well on the real unseen target domain. We train our model using the MAML algorithm on meta-train and meta-test sets. We take MDE identification loss and triplet loss as our final objective to optimize the model.

To maximize the effectiveness of the SJM module, we apply a Model-Agnostic Meta Learning (MAML) algorithm to our approach.
The MAML adopts the concept of "learning-to-learn".
It splits the source domains into meta-train and meta-test to simulate the domain bias, so as to improve the model generalization.
Our SJM module can enlarge the domain bias by enriching the style diversity of meta-train, making the learning process of learning-to-learn harder and further improving the model generalization.

Specifically, we random split the $K$ source domains into $K-1$ meta-train domains and $1$ meta-test domain.
In the meta-train stage, the images are sampled from meta-train domains and grouped into a mini-batch, denoted as $\mathcal{X}_S$.
Same as the previous training process, we fed these images into the network with the SJM module $f_{S}(\cdot;\Theta)$.
We can calculate the meta-train loss $\mathcal{L}_{mtr}$ (as same as $\mathcal{L}_{all}$) based on the stylized changed features.
% We parameterize the network as $\Theta$. 
The gradient with respect to the meta-train loss is $\nabla_{\Theta} \mathcal{L}_{mtr}(f_S(\mathcal{X}_S;\Theta))$. 
Then we optimize the network and achieve an extra updated network:
%
% \begin{align}
% \Theta' \leftarrow \Theta - \nabla_{\Theta} \mathcal{L}_{mtr}(f_S(\mathcal{X}_S;\Theta)),
% \end{align}
\begin{equation}
\begin{aligned}
\normalsize
    \mathcal{L}_{all} = \lambda \mathcal{L}_{agno} + (1 - \lambda) \mathcal{L}_{spec},
    \label{eq:all}
\end{aligned}
\end{equation}
where $\lambda$ denote the learning rate of meta-train optimizer.
In the meta-test stage, we sample images from the meta-test domain and from a mini-batch, denoted as $\mathcal{X}_T$. 
Features in meta-test stage are produced by the updated network $f(\cdot;\Theta^{'})$, where the SJM module is discarded.
The meta-test loss $\mathcal{L}_{mte}$ is calculated as well, in which $\mathcal{L}_{mte}$ is the same as $\mathcal{L}_{agno}$ because the specific-domain loss in $\mathcal{L}_{all}$ is the same as $\mathcal{L}_{agno}$ when there exists only one domain.
Finally, we utilize the combination of the meta-train and meta-test losses to optimize the original model as follows:
\begin{equation}
  \Theta \leftarrow \Theta - \gamma (\nabla_{\Theta} L_{mtr}(f_S(\mathcal{X}_S;\Theta)) + \beta \nabla_{\Theta} L_{mte}(f(\mathcal{X}_T;\Theta'))),
  \label{eq:maml-update}
\end{equation}
where $\gamma$ is the learning rate of meta-test optimizer and $\beta$ is the hyper-parameter to balance the gradient of the meta-train and meta-test losses. 
The overall training procedure of MAML is illustrated in Algorithm \ref{alg:algorithm1}.

% Because of the limited data and unlimited training time, the domain bias is easily to simulate in the original process of learning-to-learn.
% The domain bias between meta-train and meta-test domains will gradually decrease as the images of all domains are repeatedly fed into the model.
We randomly stylize images in meta-train domains and keep images in meta-test domains unchanged to mimic real train/test scenario. 
The intuition behind MAML is to expose the model to domain shift during training in hope that the model can avoid overfitting to source domain bias.
In conventional methods of applying MAML directly to the meta-train/test domain, the domain bias between meta-train/test domains will gradually decrease after the model witnesses more and more training data in the iterative process. 
% In conventional methods applying MAML directly on meta-train/test domains, the domain bias between meta-train/test domains will gradually decrease after the model has witnesses more and more training data over iterations. 
% As the model is repeatedly fed to all domain images
% the domain differences will be grad
% In the original process of learning to learn, the domain difference will be easily to simulate due to 
Whereas, our SJM module can alleviate this drawback.
More concretely, the style of features in meta-train are jittered at each iteration and the style of features in meta-test are unchanged which makes the domain gap between meta-train and meta-test domains harder to fit.
To learn a more generalizable model, the model has to ignore the influence of style variations between meta-train/test domains, and pay more attention to the identity-relevant information.

\begin{algorithm}
%\small 
\footnotesize
    \caption{Training Procedure}
    \label{alg:algorithm1}
    \KwIn{
    Source domains $ \mathcal{D}=\{\mathcal{D}_{k} \}_{k=1}^{K} $; Learning rate hyperparameters $\alpha, \beta, \gamma$.
    
    }
    \KwOut{
    Model with the SJM module $f_{S}(\cdot;{\Theta})$;
    Model w/o the SJM module $f(\cdot;{\Theta})$;
    Domain-specific classifiers $\{\varphi_{k}(\cdot; W_k)\}_{k=1}^{K}$;
    Domain-agnostic classifier $\varphi_{g}(\cdot; W_g)$.
    }
    \textbf{Initialization:} Identity style memory $\mathcal{\bar M}$; Identity similarity memory $\bar S$; Domain distance memory $\bar D$.
    
    \For{iter \textbf{in} iterations}
    %  \For{iter in iterations}
     {
        Sample $K-1$ domains as meta-train $\mathcal{D}_{mtr}$ and the remaining $\mathcal{D}_{t}$ as meta-test $\mathcal{D}_{mte}$;
        
        // For simplicity, we denote $\varphi_{k}(\cdot; W_k), k\ne t$ and $\varphi_{g}(\cdot; W_g)$ as $\varphi_{*}(\cdot; W_*)$.
        
        \textbf{Meta-training:}
        
        Sample a mini-batch $\mathcal X_{S}$ from $D_{mtr}$.
        
        Extract the normal features $F$ and jittered features $F'$ by model with the SJM module:
        \begin{align*}
            F^* = \{F;F'\} = f_{S}(X_{S};\Theta);
        \end{align*}
        
        Compute meta-train losses by Eq. \ref{eq:all}, where the memory $\mathcal{\bar M}$ is updated:
        \begin{align*}
            \mathcal L_{mtr} = \mathcal L_{all}(\varphi_{*}(F^*; W_*), F^*);
        \end{align*}
        
        Update the original model and meta-train classifiers parameters by:
        \begin{align*}
            {\Theta}' &\leftarrow \Theta - \alpha \nabla_{\Theta}  \mathcal L_{mtr}; \\
            {W_{*}'} &\leftarrow W_{*}' - \alpha \nabla_{W_{*}}  \mathcal L_{mtr};
        \end{align*}
        
        Calculate similarity $S$ with global weight ${W_g}'$ by Eq. \ref{eq:similarity} and update the memory $\bar S$;
        
        Calculate domain distances with extracted features $F^*$ by Eq. \ref{eq:mmd} and update the memory $\bar D$;
        
        \textbf{Meta-testing:}
        
        Sample a mini-batch $\mathcal X_{T}$ from $D_{mte}$.
        
        Extract the normal features $F$ by model w/o the SJM module::
        \begin{align*}
            F = f(X_{T};\Theta');
        \end{align*}
        
        Compute meta-test losses w/o the SJM module by Eq. \ref{eq:all}: 
        \begin{align*}
            \mathcal L_{mte} = \mathcal L_{all}(\varphi_{t}(F; W_t), F)
        \end{align*}
        
        Update the meta-test classifier parameters by:
        \begin{align*}
            {W_{t}'} &\leftarrow W_{t}' - \beta \nabla_{W_{t}}  \mathcal L_{mte};
        \end{align*}
        
        \textbf{Meta-optimizing}
        
        Update the original model parameters $\Theta$ by:
        \begin{align*}
            \Theta \leftarrow \Theta - \gamma (\nabla_{\Theta} \mathcal L_{mtr}
            + \beta \nabla_{\Theta} \mathcal L_{mte})
        \end{align*}
    }

\end{algorithm}
\section{Experiments}
\label{sec:experiment}
% See \Table{sota} and \Table{ablations}.

\subsection{Datasets and Evaluation Metrics}

% 需要强调一下 dataset 的摄像头，为 single-domain 划分做准备；

Following the previous works \cite{jia2019frustratingly,song2019generalizable,jin2020style,dai2021generalizable}, we conduct our experiments on public person ReID or Pearson-search datasets, including Market1501 \cite{zheng2015scalable}, DukeMTMC-reID \cite{zheng2017unlabeled}, CUHK02 \cite{li2013locally}, CUHK03 \cite{li2014deepreid}, MSMT17 \cite{wei2018person}, CUHK-SYSU \cite{xiao2017joint} and four small ReID datasets including PRID \cite{hirzer2011person}, GRID \cite{loy2010time}, VIPeR \cite{gray2008viewpoint}, and iLIDs \cite{zheng2009associating}.
\textbf{Multi-source DG ReID task.}
On multi-source DG ReID task, we evaluate our methods on large-scale datasets and small-scale datasets, respectively. 
On large-scale datasets, we train and evaluate on four large datasets, \textit{i.e.} Market-1501, DukeMTMC, CUHK03 and MSMT17. Concretely, we choose one of these four datasets as evaluation dataset and use the remaining three datasets as our training datasets. 
Different from evaluation strategies on large datasets, two evaluation protocols (termed as protocol-1 and protocol-2) are adopted on small-scale datasets for the comparison with previous methods.
Both two protocols are evaluated on all small-scale datasets (\textit{i.e.} PRID, GRID, VIPeR and iLIDs) and trained on different datasets.
Specifically, in protocol-1, we train our model on six large-scale datasets, \textit{i.e.} Market-1501, DukeMTMC, CUHK02, CUHK03, CUHK-SYSU and MSMT17. 
In protocol-2, we train our methods on four datasets, \textit{i.e.} CUHK03, DukeMTMC, Market-1501 and MSMT17. 

\textbf{Single-source DG ReID task.}
We have also conduct experiments on single-source ReID DG task. We use two datasets Market-1501 and DukeMTMC in our experiments. In the single-source setting, one of these datasets is used for training and the other one is used for evaluation.

Following the common evaluation metric, the cumulative matching characteristic (CMC) at Rank-$k$ and mean average precision (mAP) are used to evaluate the model's performance on target domains.

\begin{table*}[ht]
\tabcolsep=4.5pt
\small
\centering
  \caption{Comparison (\%) with the state-of-the-arts DG ReID methods on four large-scale person ReID bencnmarks.
	}
  \begin{tabular}{lccccccccccccc}
 \hline
	\multirow{2}{*}{Method} &
	\multirow{2}{*}{Backbone} & 
	\multirow{2}{*}{Reference} &  \multicolumn{2}{c}{D+M+MS$\rightarrow$C3} & \multicolumn{2}{c}{C3+M+MS$\rightarrow$D} & \multicolumn{2}{c}{C3+D+MS$\rightarrow$M} & \multicolumn{2}{c}{C3+D+M$\rightarrow$MS}  & \multicolumn{2}{c}{Average}\\ 
	\cline{4-13}
	& & & mAP & Rank-1 & mAP & Rank-1 & mAP & Rank-1 & mAP & Rank-1 & mAP & Rank-1 \\ \hline
  % DEMO $^*$ & \multirow{10}{*}{C2+C3+D+M+CS} & TIP 2019 & - & - & - & - & - & - & - & -  \\
  QAConv \cite{liao2020interpretable} & ResNet50 & ECCV 2020 & 21.00 & 23.50 & 47.10 & 66.10 & 35.60 & 65.70 & 7.50 & 24.30  & 27.80 & 44.90 \\
  OSNet \cite{zhou2019omni} & OSNet & ICCV 2019 & 23.30 & 23.90 & 47.00 & 65.20 & 44.20 & 72.50 & 12.60 & 33.20  & 31.77 & 48.70 \\
  SNR \cite{jin2020style} & ResNet50 &  CVPR 2020 & 29.00 & 29.10 & 48.30 & 66.70 & 48.50 & 75.20 & 13.80 & 35.10  & 34.90 & 51.52 \\
  M$^3$L \cite{zhao2021learning} & ResNet50 & CVPR 2021 & 29.90 & 30.70 & 50.30 & 69.40 & 48.10 & 74.50 & 12.90 & 33.00  & 35.30 & 51.90 \\
  MECL \cite{yu2021multiple} & ResNet50 & Arxiv & 31.50 & 32.10 & 53.40 & 70.00 & 56.50 & 88.00 & 13.30 & 32.70  & 38.68 & 55.70 \\
  RaMoE \cite{dai2021generalizable} & ResNet50 & CVPR 2021 & 35.50 & 36.60 & 56.90 & 73.60 & 56.50 & 82.00 & 13.50 & 34.10  & 40.60 & 56.57 \\
  Baseline & ResNet50 & - & 32.60 & 32.90 & 49.40 & 65.80 & 49.90 & 75.40 & 9.90 & 13.50  & 35.45 & 46.90 \\
  SVIL (Ours) & ResNet50 & - & \textbf{38.50} & \textbf{37.82} & \textbf{58.13} & \textbf{74.78} & \textbf{59.21} & \textbf{82.36} & \textbf{17.06} & \textbf{39.99}  & \textbf{43.23} & \textbf{58.74} \\
  \hline
  M$^3$L \cite{zhao2021learning} & IBN-ResNet50 & CVPR 2021 & 32.10 & 33.10 & 51.10 & 69.20 & 50.20 & 75.90 & 14.70 & 36.90  & 37.02 & 53.78 \\
  MECL \cite{yu2021multiple} & IBN-ResNet50 & Arxiv & 37.30 & 38.10 & 57.20 & 74.10 & 60.90 & 83.20 & 18.00 & 41.20  & 43.35 & 59.15 \\
  SVIL (Ours) & IBN-ResNet50 & - & \textbf{43.22} & \textbf{44.07} & \textbf{61.68} & \textbf{77.33} & \textbf{64.95} & \textbf{86.05} & \textbf{19.85} & \textbf{44.63}  & \textbf{47.43} & \textbf{63.02} \\
 \hline
  \end{tabular}
  \label{tab:sota_4}
\end{table*}
\subsection{Implementation Details}
We adopt ResNet50 \cite{ResNet} and IBN-ResNet50 \cite{IBNNet} pretrained on ImageNet as our backbones, respectively. 
% Baseline 配置
We add a batch normalization (BN) layer after the global pooling layer to get the ReID feature. 
A linear classifier is added after BN layer to get classification predictions used for calculating cross-entropy loss and triplet loss \cite{hermans2017defense}. 
We use the above structure as our baseline. Note that this baseline is different from the strong baseline we proposed.
Images are resized to $256\times 128$, and random cropping and random flipping are utilized as data augmentation. 
% We perform random cropping and random flipping for data aon training images. 
The batch size of each specific domain is set to 128, including 32 identities and 4 images per identity.
% We set the batch size to 64, including 16 identities and 4 images per identity.
In training stage, we evenly sample mini-batch from each source domain and combine all these batches as model's input. 
For optimizing the model, we use Adam optimizer with a weight decay of $5\times 10^{-4}$. 
The learning rate of meta-train phase and meta-test phase are initialized as $3.5\times 10^{-4}$ and are decayed by 0.1 at the 30th and 50th epochs respectively.
We train the model on 4 GTX 1080Ti GPUs for 90 epochs.
% 缺少Meta learning 的配置
Our SJM module is plugged after the stage-1 of backbone, and the its scale $S$ is set to 4.
In addition, the $\lambda$ in $\mathcal{L}_{all}$ is 0.1.

\subsection{Comparison with State-of-the-Art methods}
We compare our methods with the state-of-the-art methods on the multi-source DG ReID task on both large-scale datasets and small-scale datasets. We have also conduct experiments on single-source DG ReID task to demonstrate the effectiveness and generalization ability of our methods.
% \input{tab/sota_5}

% \textbf{Results on small-scale ReID datasets}
% There are two evaluation settings for DG-ReID on small-scale ReID datasets. One is to train the model on C2+C3+D+M+CS, and the other one is to train the model on C3+D+M+MS. Both these two settings are to evaluate the model on four small-scale ReID datasets: PRID, GRID, VIPeR, iLIDs respectively. We compare our proposed method with the state-of-the-arts (SOTAs) under these two settings. For the fair of comparison, we provide both results based on ResNet-50 and IBN-50. 
%
% 结合实验数据说明本方法可以超过所有的 SOTA 方法
%

\subsubsection{Results on large ReID datasets under multi-source tasks}
% \textbf{Results on large-scale ReID datasets}
% When evaluated on the large-scale ReID datasets.
% \subsubsection{Results on large ReID datasets}
% \subsubsection{Results on large-scale ReID datasets}
To demonstrate the effectiveness of our method, we compare it with the state-of-the-arts (SOTAs) on the large-scale DG ReID benchmark, including QAConv \cite{liao2020interpretable}, SNR \cite{jin2020style}, M3L \cite{zhao2021learning}, RaMoE \cite{dai2021generalizable} and OSNet \cite{zhou2019omni}. 
For simplicity, we denote Market-1501 as M, DukeMTMC-reID as D, CUHK03 as C3, MSMT17 as MS.

% \textbf{Multi-source DG ReID task.}
Following the evaluation protocol for DG ReID in \cite{dai2021generalizable,zhao2021learning},
we utilize the leave-one-out protocol to split these four datasets into training/testing datasets. 
Specifically, we select three datasets as multiple training datasets and the remaining one is used as test dataset. 
All images in training datasets are utilized for training regardless of train/test splits.
% 这一段的比较需要再补充点，太少了
The results are shown in Tab \ref{tab:sota_4}. 
For the fair of comparison, we provide both results based on ResNet50 and IBN-ResNet50.
Under the ResNet50 backbone, our SVIL outperforms our baseline by 9.31\%, 8.73\%, 5.90\% and 7.16\% in mAP.
Under the ResNet50 backbone, our SVIL outperforms the second best method by 2.71\%, 1.23\%, 3.00\% and 3.26\% on M, D, C and MS in mAP respectively. 
Under the IBN-ResNet50 backbone, our method outperforms the second best method by 4.05\%, 4.48\%, 5.92\% and 1.85\% in mAP.
The above experiments show our method can achieve state-of-the-art performance on all target domains under different backbones and can outperform baseline by a large margin, which demonstrates the effectiveness and superiority of our method.

\begin{table*}[ht]
\tabcolsep=6pt
\small
\centering
  \caption{Comparison (\%) with the state-of-the-arts DG ReID methods on small-scale person ReID bencnmarks.
	}
  \begin{tabular}{lccccccccccccc}
 \hline
	\multirow{2}{*}{Method} & \multirow{2}{*}{Source} & \multicolumn{2}{c}{Target: PRID} & \multicolumn{2}{c}{Target: GRID} & \multicolumn{2}{c}{Target: VIPeR} & \multicolumn{2}{c}{Target: iLIDs} & \multicolumn{2}{c}{Average} \\ 
	\cline{3-12}
	& & mAP & Rank-1 & mAP & Rank-1 & mAP & Rank-1 & mAP & Rank-1 & mAP & Rank-1 \\ \hline
  % DEMO $^*$ & \multirow{10}{*}{C2+C3+D+M+CS} & TIP 2019 & - & - & - & - & - & - & - & -  \\
  DIMN \cite{song2019generalizable} & \multirow{10}{*}{C2+C3+D+M+CS}  & 51.95 & 39.20 & 41.09 & 29.28 & 60.12 & 51.23 & 78.39 & 70.17 & 57.89 & 47.47 \\
  DualNorm \cite{jia2019frustratingly} & & 64.90 & 60.40 & 45.70 & 41.40 & 58.00 & 53.90 & 78.50 & 74.80 & 61.78 & 57.62 \\
  MMD-AAE \cite{li2018domain} & & - & 57.20 & - & 47.40 & - & 58.40 & - & 84.80 & - & 61.95   \\
  SNR \cite{jin2020style} & & 66.50 & 52.10 & 47.70 & 40.20 & 61.30 & 52.90 & 89.90 & 84.10 & 66.35 & 57.33 \\
  RaMoE \cite{dai2021generalizable} & & 67.30 & 57.70 & 54.20 & 46.80 & 64.60 & 56.60 & \textbf{90.20} & 85.00 & 69.08 & 61.52 \\
  %MetaBINs & & \textbf{81.00} & \textbf{74.20} & 57.90 & 48.40 & 68.60 & 59.90 & 87.00 & 81.30 & 73.62 & 65.95 \\
  Baseline  & & 63.77 & 53.00 & 50.63 & 39.00 & 62.76 & 53.23 & 82.34 & 76.67 & 64.88 & 55.47 \\
%  Strong Baseline  & & 67.77 & 57.00 & 54.63 & 43.00 & 66.76 & 57.23 & 86.34 & 80.67 & 68.88 & 59.47 \\
  SVIL (Ours) & & \textbf{69.39} & \textbf{58.05} & \textbf{56.18} & \textbf{44.20} & \textbf{71.70} & \textbf{63.29} & 88.97 & \textbf{85.30} & \textbf{71.56} & \textbf{62.63} \\
 \hline
  SNR \cite{jin2020style} & \multirow{6}{*}{C3+D+M+MS} & 60.00 & 49.00 & 41.30 & 30.40 & 65.00 & 55.10 & 91.90 & 87.00 & 64.55 & 55.38 \\
  RaMoE \cite{dai2021generalizable} & & 66.80 & 56.90 & 53.90 & 43.40 & 72.70 & 63.40 & \textbf{92.30} & \textbf{88.40} & 71.42 & 63.02 \\
  Baseline & & 65.18 & 55.00 & 52.88 & 42.60 & 70.36 & 62.19 & 86.03 & 82.00 & 68.74 & 60.45 \\
%   Strong Baseline & & 65.18 & 55.00 & 55.88 & 44.60 & 72.36 & 64.19 & 88.03 & 84.00 & 70.49 & 61.95 \\
  SVIL (Ours) & & \textbf{75.28} & \textbf{67.00} & \textbf{56.90} & \textbf{48.00} & \textbf{75.30} & \textbf{67.09} & {90.03} & {86.67} & \textbf{74.38} & \textbf{67.19} \\
 \hline
  \end{tabular}
  \label{tab:sota_5}
\end{table*}
\subsubsection{Results on small ReID datasets under multi-source tasks}
Following the previous works \cite{jia2019frustratingly,song2019generalizable,jin2020style,dai2021generalizable}, we adopt the Protocol-1 \cite{song2019generalizable} and Protocol-2 \cite{jin2020style} to evaluate our method on four small ReID datasets, respectively.
For simplicity, we also denote CUHK02 as C2 and CUHK-SYSU as CS.
% Three exist three different evaluation protocols as shown in [TAB]. 
Under Protocol-1, all images in M+D+C2+C3+CS are utilized for training regardless of train/test splits and the trained network was tested on PRID, GRID, VIPeR and iLIDs, repectively.
Our method is higher than other methods on PRID, GRID and VIPeR datasets under the mAP and Rank-1 evaluation metric.
In addition, the proposed method achieves greater improvement than baseline on all datasets.
Under Protocol-2, all images in M+D+C3+MS are utilized for training regardless of train/test splits.
Our approach outperforms the baseline by a large margin and obtains competitive results compared to SOTAs.

% Under Protocol-1 and Protocol-2, method was tested on PRID, GRID, VIPeR and iLIDs repectively.  Protocol-3 is consisted of M+D+C+MS, and under it three domains are used for training (including the training and testing images) and the remaining one is used for testing. Previous works were evaluated on different protocols. To compare with them, we evaluate our method on all three evaluation protocols.

% 补一个test的描述

% \textbf{Results in single domain.}

\begin{table}[ht]
\small
\centering
\caption{Performance (\%) comparison with the state-of-the-arts on the single-source DG-ReID task.
	}
  \begin{tabular}{c|cc|cc}
  \hline
	\multirow{2}{*}{Method} & \multicolumn{2}{c|}{M$\rightarrow$D} & \multicolumn{2}{c}{D$\rightarrow$M} \\ 
	\cline{2-5}
	& mAP & Rank-1 & mAP & Rank-1 \\ \hline
  % DEMO $^*$ & \multirow{10}{*}{C2+C3+D+M+CS} & TIP 2019 & - & - & - & - & - & - & - & -  \\
  IBN-ResNet \cite{IBNNet} & 24.30 & 43.70 & 23.50 & 24.00 \\
  OSNet \cite{Zhou2021LearningGO} & 25.90 & 44.70 & 24.00 & 52.20 \\
  OSNet-IBN \cite{Zhou2021LearningGO} & 27.60 & 47.90 & 27.40 & 57.80 \\
  CrossGrad \cite{Shankar2018CrossGard} & 27.10 & 48.50 & 26.30 & 56.70 \\
  QAConv \cite{Liao2019QAConv} & 28.70 & 48.80 & 27.20 & 58.60 \\
  L2A-OT \cite{Zhou2020L2G} & 29.20 & 50.10 & 30.20 & 63.80 \\
  OSNet-AIN \cite{Zhou2021LearningGO} & 30.50 & 52.40 & 30.60 & 61.00 \\
  SNR \cite{jin2020style} & 33.60 & 55.10 & 33.90 & 66.70 \\
  MetaBIN \cite{choi2021meta} & 33.10 & 55.20 & \textbf{35.90} & \textbf{69.20} \\
  SVIL (Ours) & \textbf{35.08} & \textbf{56.01} & 35.05 & 67.91 \\
  \hline
  \end{tabular}
  \label{tab:single}
\end{table}
% \textbf{Effectiveness of our method on single source task.}
% \textbf{Single-source DG-ReID task.}
% To demonstrate the effectiveness of our methods, we conduct experiments of single-source DG-ReID task on Market1501 (M) \cite{zheng2015scalable}, and DukeMTMC-reID (D) \cite{zheng2017unlabeled} datasets.
% We conduct experiments of single source task on Market1501 and DukeMTMC.
\subsubsection{Results on large ReID datasets under multi-source tasks}
Since the images of the same dataset come from different cameras, which generally have different styles as mentioned in Sec. \ref{sec:intro}, we divide the single source dataset into $K$ sub-datesets according to the camera information to simulate the setting of multiple source.
More specifically, Market1501 \cite{zheng2015scalable}, DukeMTMC-reID \cite{zheng2017unlabeled} datasets are divided into $3$ subsets according to their cameras, where the camera labels from different subsets do not conflict with each other.
Through the division, style variations appear in various sub-datasets.
To keep it simple, we discard the domain-specific loss $L_{spec}$ and preserve the domain-agnostic loss $L_{agno}$ with the $\lambda = 1$.
% We first divide the single-source dataset into multiple sub-datasets to simulate the setting of multi-source DG-ReID task as described in Sec \ref{sec:moreDetails}.
We then embed our method into the single-source DG-ReID task in the same way as multi-source task.
`M $\rightarrow$ D' in Tab. \ref{tab:single} indicates that Market1501 is the labeled source training domain and DukeMTMC-reID is the unseen target domain.
Experimental results show that our approach achieves competitive performances with SOTAs.
% Experimental results show that, though the domain-specific loss is discarded, our approach achieves competitive performances with SOTAs.
However, there is still room for improvement in applying our method to single-domain problems.
On the one hand, our division of single source domain is rough.
When the subsets are divided only according to the camera label, the identity label space of each subset would overlaps, which is inconsistent with the real multi-source DG-ReID problem.
On the other hand, the domain-specific loss is discarded.
Both of these factors will prevent our method from achieving better performance.
Nonetheless, the comparison in Tab. \ref{tab:single} also proves that our method can improve the generalization capability of model.

\begin{table}[t]
% 这个图可能需要加一列 No. x 序号
\tabcolsep=1.5pt
\small
\centering

\caption{Ablations studies on different components of our method.
	}
  \begin{tabular}{l|ccc|cc|cc}
  \hline
	\multirow{2}{*}{Backbone} & \multirow{2}{*}{SJM} & \multirow{2}{*}{MAML} & \multirow{2}{*}{Loss}  & \multicolumn{2}{c|}{C+D+MS$\rightarrow$M} & \multicolumn{2}{c}{C+M+MS$\rightarrow$D} \\ 
	\cline{5-8}
	& & & & mAP & Rank-1 & mAP & Rank-1 \\ \hline
  % DEMO $^*$ & \multirow{10}{*}{C2+C3+D+M+CS} & TIP 2019 & - & - & - & - & - & - & - & -  \\
  ResNet50 & $\times$ & $\times$ & $\mathcal{L}_{spec}$ & 50.74 & 77.32 & 52.64 & 69.52 \\
  ResNet50 & \checkmark & $\times$ & $\mathcal{L}_{spec}$ & 53.76 & 80.40 & 53.99 & 71.50 \\
  IBN-ResNet50 & $\times$ & $\times$ & $\mathcal{L}_{spec}$ & 58.12 & 81.80 & 57.87 & 73.43 \\
  IBN-ResNet50 & \checkmark & $\times$ & $\mathcal{L}_{spec}$ & 58.77 & 82.36 & 58.39 & 73.83 \\
  \hline 
  ResNet50 & $\times$ & \checkmark & $\mathcal{L}_{spec}$ & 56.09 & 80.29 & 55.74 & 71.23 \\
  ResNet50 & \checkmark & \checkmark & $\mathcal{L}_{spec}$ & 58.18 & 82.33 & 57.26 & 73.61 \\
  IBN-ResNet50 & $\times$ & \checkmark & $\mathcal{L}_{spec}$ & 61.40 & 83.49 & 60.24 & 75.58 \\
  IBN-ResNet50 & \checkmark & \checkmark & $\mathcal{L}_{spec}$ & 63.95 & 84.77 & 60.86 & 76.26 \\
  \hline
  IBN-ResNet50 & $\times$ & \checkmark & $\mathcal{L}_{all}$ & 62.30 & 83.43 & 60.72 & 76.26 \\
  IBN-ResNet50 & \checkmark & \checkmark & $\mathcal{L}_{all}$ & 64.95 & 86.05 & 61.68 & 77.33 \\
  \hline
  \end{tabular}
  \label{tab:abalation}
\end{table}

\begin{table}[t]
% 这个图可能需要加一列 No. x 序号
\tabcolsep=2pt
\small
\centering

\caption{Ablations studies of the SJM Module.
	}
  \begin{tabular}{l|c|cc|cc}
  \hline
	\multirow{2}{*}{Method}  & \multirow{2}{*}{Cross-domain}  & \multicolumn{2}{c|}{C+D+MS$\rightarrow$M} & \multicolumn{2}{c}{C+M+MS$\rightarrow$D} \\ 
	\cline{3-6}
	& & mAP & Rank-1 & mAP & Rank-1 \\ \hline
  % DEMO $^*$ & \multirow{10}{*}{C2+C3+D+M+CS} & TIP 2019 & - & - & - & - & - & - & - & -  \\
  SVIL w/o SJM & $\times$ & 62.30 & 83.43 & 60.72 & 76.26 \\
  SVIL w/ SJM & $\times$ & 64.01 & 85.34 & 61.38 & 76.91 \\
  SVIL w/ SJM & \checkmark & 64.95 & 86.05 & 61.68 & 77.33 \\
  \hline
  \end{tabular}
  \label{tab:sjm_aba}
\end{table}

\begin{table}[t]
\small
\centering
\caption{Study about which stage of IBN-ResNet50 to plug the SRJ module.
	}
  \begin{tabular}{c|cc|cc}
  \hline
	\multirow{2}{*}{Method} & \multicolumn{2}{c|}{C+D+MS$\rightarrow$M} & \multicolumn{2}{c}{C+M+MS$\rightarrow$D} \\ 
	\cline{2-5}
	& mAP & Rank-1 & mAP & Rank-1 \\ \hline
  % DEMO $^*$ & \multirow{10}{*}{C2+C3+D+M+CS} & TIP 2019 & - & - & - & - & - & - & - & -  \\
  SVIL w/o SJM & 62.30 & 83.43 & 60.72 & 76.26 \\
  \hline
  SJM after stage-0 & 64.44 & 85.15 & 61.16 & 77.06 \\
  SJM after stage-1 & 64.95 & 86.05 & 61.68 & 77.33 \\
  SJM after stage-2 & 62.90 & 84.83 & 59.13 & 74.46 \\
  SJM after stage-3 & 58.34 & 81.41 & 59.01 & 75.49 \\
  SJM after stage-4 & 50.62 & 75.00 & 51.60 & 70.11 \\
  \hline
  \end{tabular}
  \label{tab:stage}
\end{table}

\subsection{Ablation Study}

% \paragraph{Comparison of Backbone Settings}
\textbf{Effectiveness of our SJM module on different backbones.}
We plug the SJM module into the two backbones, ResNet-50 \cite{ResNet} and IBN-ResNet-50 \cite{IBNNet}, and conduct the experiments with different train strategy.
% under two settings of the training strategy with/without meta learning, respectively.
As shown in Tab. \ref{tab:abalation}, the results of methods with SJM are all higher than the methods w/o SJM.
For instance, mAP/Rank-1 of ResNet50+SJM is 3.02\%/3.08\% higher than ResNet50, and mAP/Rank-1 of IBN-ResNet50+SJM is 0.65\%/0.56\% higher than IBN-ResNet50 on M target domain.
There is less improvement in IBN-ResNet50 because the Instance Normalization (IN) in IBN-ResNet50 can improve the style-insensitive of the model as well.
Its effects are partially repeated with our
SJM module.
% The effects of the SJM and IN are partially repeated.
In addition, the SJM also perform well on the ResNet50/IBN-ResNet50 with MAML.
These comparisons confirm the effectiveness of our SJM module.
% When not using MAML, the SJM bring an improvement of +3.02\%/+3.08\% 
% As shown in Tab. \ref{tab:abalation}, the method in ResNet50 w/o MAML 
% We have inserted the c module into the two backbones of a and b with and without the meta learning training strategy.

\textbf{Effectiveness of our SJM module with MAML.}
We conducted experiments on two backbones (\textit{i.e.} ResNet50 \cite{ResNet} and IBN-ResNet50 \cite{IBNNet} ) to prove the effectiveness of the combination of the SJM module and MAML.
As shown in Tab. \ref{tab:abalation}, the improvement of SJM module on network with MAML is great than the improvement on network w/o MAML.
Take C+D+MS$\rightarrow$M as an example, mAP/Rank-1 of IBN-ResNet50+SJM+MAML is 2.55\%/1.28\% higher than IBN-ResNet50+MAML,
and mAP/Rank-1 of IBN-ResNet50+SJM is 0.65\%/0.56\% higher than IBN-ResNet50.
% 哪个效果更好写在 x 上
The comparison demonstrates that the combination of the SJM module and MAML would produce a synergy effect and achieves better performance.
% Both results prove that the combination of a and b will produce a synergistic effect and achieve better performance
% Our module is very compatible with maml, which brings greater performance improvement
% Synergy effect 协同效应，商业/化学用语，形容 1+1 >2

\textbf{Effectiveness of our SJM module on different loss functions.}
% \textcolor{red}{There are three losses in our method, \textit{i.e.}, $\mathcal{L}_{spec}$, $\mathcal{L}_{agno}$ and $\mathcal{L}_{all}$ ($\mathcal{L}_{spec}$ + $\lambda \mathcal{L}_{agno}$), where the samples of the sample pairs in $\mathcal{L}_{agno}$ may come from different domains.
% As described in Sec. \ref{sec:SJM}, the SJM module can make cross-domain samples harder and prompt the model to focus on the identity-related discriminative information.
% Results in Tab. \ref{tab:abalation} confirm the above-mentioned effect of our SJM module.}
% The experimental results also prove the above-mentioned effect of smm.
The results in Table \ref{tab:abalation} demonstrate the effectiveness of the SJM module with various losses.
For instance, in C+D+MS$\rightarrow$M, the improvement of $\mathcal{L}_{agno}$ on IBN-ResNet-50+MAML+$\mathcal{L}_{spec}$+SJM is 0.1\%/1.37\% higher than the improvement on IBN-ResNet-50+MAML+$\mathcal{L}_{spec}$ on mAP/Rank-1 metrics.
This comparison confirms that our SJM module can make cross-domain samples harder.
% that  achieve a greater improvement in the model with the SJM module.
%\mathcal{L}_{all} = \mathcal{L}_{spec} + \lambda \mathcal{L}_{agno},
% \paragraph{Capability of single domain generalization}

\textbf{Effect of different identity relationship modeling.}
We conduct experiments to confirm the effectiveness of identiy relationship modeling in the SJM module.
The results in Table \ref{tab:sjm_aba} show that only hard identity emphasis can improve the results, and cross-domain identity emphasis can further improve the model performance on the basis.

\begin{figure}
\begin{center}
\includegraphics[width=1.\columnwidth]{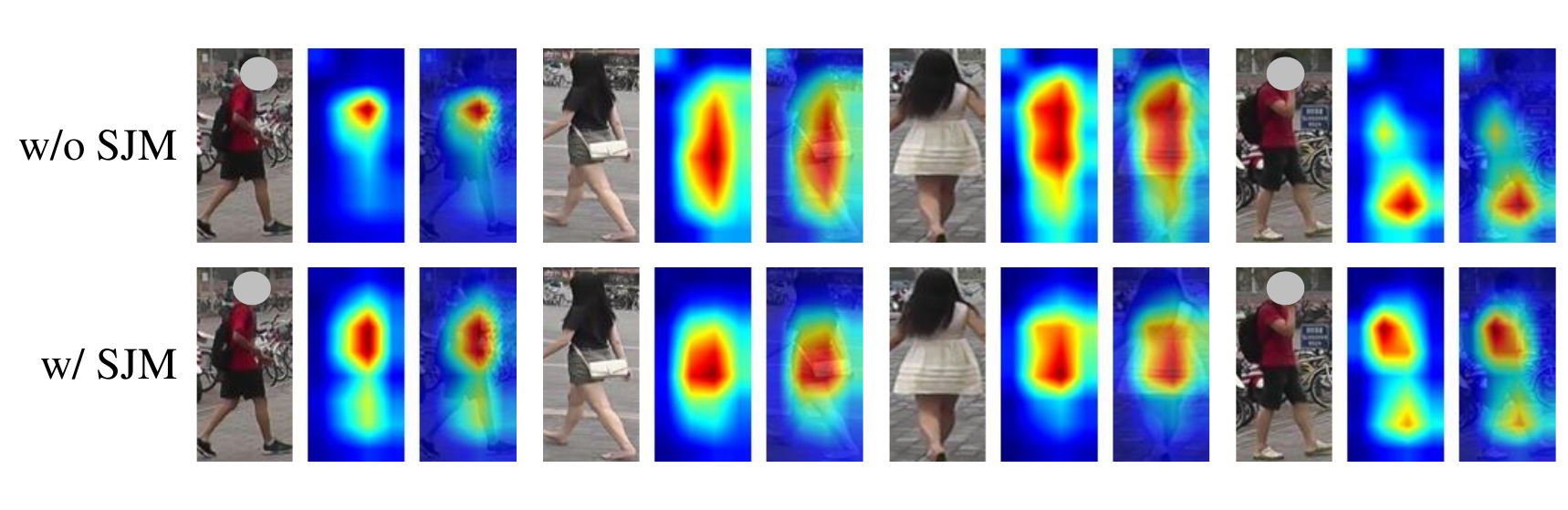}
\end{center}
\caption{
Visualization of activation maps of different features on method w/o SJM and method w/ SJM.
The maps of method with SJM pay more attention to identity-relevant discriminative information, \textit{e.g.} clothes and bags.
}
\label{fig:featmap}
\end{figure}
\begin{figure*}
\begin{center}
\includegraphics[width=2.\columnwidth]{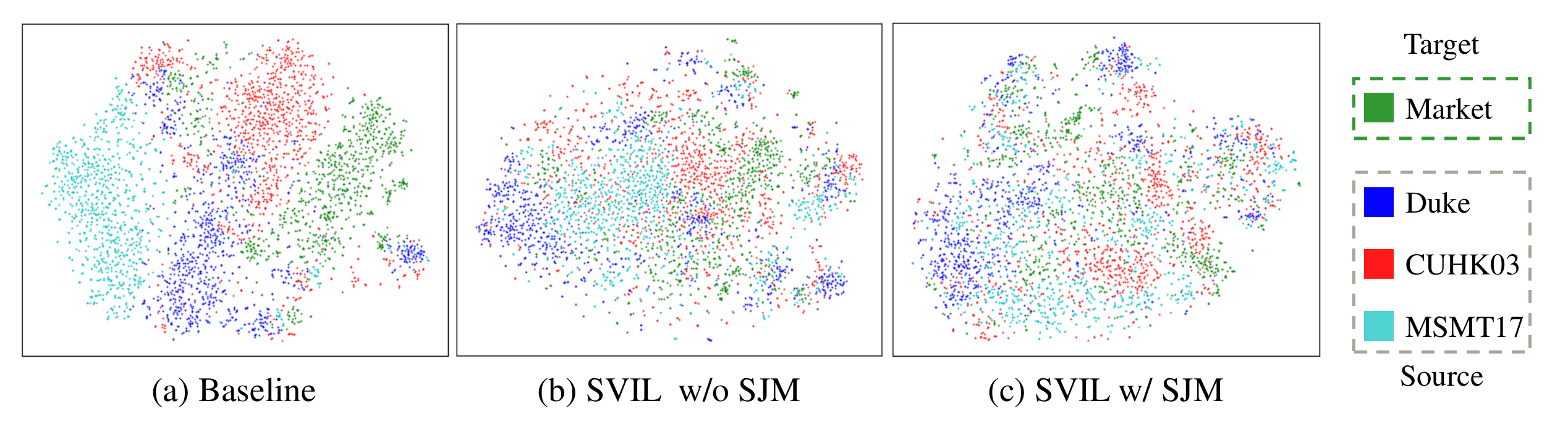}
\end{center}
\caption{
Visual distributions of four person ReID benchmarks.
The distributions are achieved from inference features of (a) Baseline, (b) Our method without SJM module and (b) Our method with SJM module.
All of the backbones are ResNet-50 and the dimension of inference features is reduced by t-SNE \cite{van2008visualizing}.
}
\label{fig:scatter}
\end{figure*}
\begin{figure*}[ht]
\begin{center}
% 后续再换 pdf 格式
\includegraphics[width=2.\columnwidth]{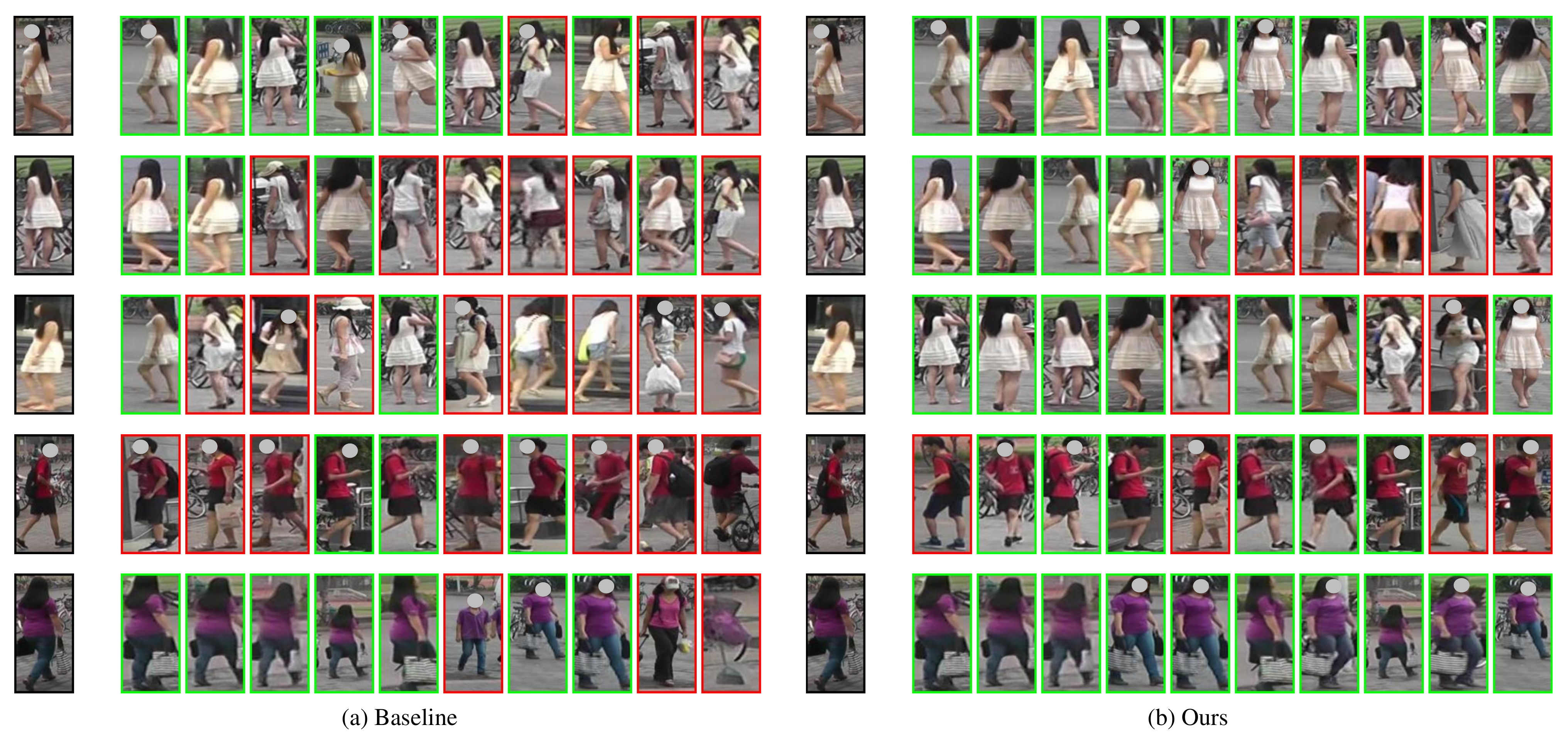}
\end{center}
\caption{
Examples of ranking results on Market1501. The green and red boxes indicate the correct matchings and the wrong matchings, respectively. (a) is the results of baseline, and (b) is the results of our methods.
}
\label{fig:rank}
\end{figure*}
% \paragraph{Which Stage to Add the SMM?}
\textbf{Effectiveness of which stage to plug the SJM Module.}
Our SJM module is a plug-and-play module which can be plugged after any stage of the backbone network.
We conduct experiments on the Strong baseline with IBN-ResNet backbone, which has five stages: the stage-0 consists of Conv, BN and Max Pooling layer, and stage-1/2/3/4 are the other four convolutions blocks.
We plug the SJM module after different stages to study how different depth will affect the performance of SJM module.
Results in Tab. \ref{tab:stage} shows that our method performs better in shallow stages (0, 1) and worse in deeper stages (2, 3, 4).
This phenomenon is because that shallow features contain more low-level features, and the style transformation on them would lose less semantic information.
% 这一块需要那个结论：浅层特征的语义特征弱，风格的变换可以带来背景等非identity信息的变化；
% 而深层的网络语义信息强，风格信息少，风格的抖动会引起 identity 的丢失。
% Then we plug the SJM module after stage-1 in all our experiments if not specified.

% \input{tab/scale}
% % \paragraph{Effectiveness of multi-scale SMM}
% \textbf{Effectiveness of the style scale in our SJM module.}
% The scale $S$ determines the granularity of the local stylized feature.
% When $S=1$, the local stylized feature becomes a global one.
% The Tab. \ref{tab:scale} shows the results of our method with different scale.
% As $S$ increases, the performance improves at first.
% Whereas, performances does not always increase with $S$.
% It achieves the best results in $S=4$.
% This may be because that, when $S$ is too large, the style feature will focus on the subtle but trivial local information, resulting in a lack of useful style information.

\subsection{Qualitative Analysis}

% \begin{table}[t]
% \tabcolsep=4pt
% \small
% \centering
%   \caption{mAP(\%) of method trained by various datasets.
% 	}
%   \begin{tabular}{lccc}
%  \hline
%  Method & D$\rightarrow$M & D+C$\rightarrow$M & D+C+MS$\rightarrow$M \\
%  \hline
%  ResNet & 28.09 & 43.33 & 50.74 \\
%  ResNet + SJM & 30.79(+2.70) & 46.06(+2.73) & 53.76(+3.02) \\
%  \hline
%   \end{tabular}
%   \label{tab:ms}
% \end{table}

% \textbf{Affect of the number of source domain.} We train ResNet and +SJM with variant number of source datasets.
% Results in Tab \ref{tab:ms} show our SJM can perform better when the number of source domains increases, which confirm the effectiveness of our method.

% \paragraph{Feature Map Visualization.}
\textbf{Visualization of Feature Maps.}
To better understand the influence of our SJM module, we visualize the intermediate feature maps in the method w/o SJM and method w/ SJM in Fig. \ref{fig:featmap}.
Following \cite{jin2020style}, we obtain each activation map by summarizing the feature maps along channels followed by a spatial $l_2$ normalization.
% Fig. \ref{fig:featmap} shows the activation map of images from the strong baseline and our method, respectively.
We can observe that the maps of our method pay more attention on discriminate areas.
For instance, as shown in the 2nd and 3rd columns in the maps, method w/ SJM tends to focus more on bags and skirts, which have some discriminative details.
Whereas the focus area of the method w/o SJM is scattered, including part of background and other unimportant areas.
These areas may contain more style information.
% This phenomenon confirms the effectiveness of our methods.
This phenomenon confirms that our SJM module contributes to learning a more style-insensitive and more generalized model.

% \paragraph{Distributions of Features Visualization.}

\textbf{Visualization of Feature Distributions.}
In Fig. \ref{fig:scatter}, we visualize the t-SNE \cite{van2008visualizing} distributions of the features on the four benchmarks for baseline and our method.
Different colors represent various datesets (\textit{i.e.,} bule, red and cyan denote the source datasets Duke, CUHK03 and MSMT17, respectively; green indicates the target dataset Market1051.).
As shown in Fig. \ref{fig:scatter} (a), we can see the feature distributions of different datasets on baseline are separately distributed and have visible domain gaps.
After adopting our method, these domain gaps are closed a lot.
The features of different domains are mixed together.

\textbf{Visualization of Rank List.}
% 写的太少了，加一些
Fig. \ref{fig:rank} demonstrates 5 pairs of ranking results, \textit{i.e.}, the ranking results of baseline in (a) and the ranking results of our method in (b).
We can observe that the wrong matching in baseline generally have similar clothing or background, which share similar styles.
Our method can reduce the mismatches caused by style changes.
This phenomenon confirms the effectiveness of our approach.

% \section{Discussion of Limitations}
% % Our method adopt MAML as our training scheme, the process of copying the model and calculate second-order gradients consumes additional computation resources and cause a longer training time.
% The current framework of our method requires multiple source domains, which can not be applied to the single domain generalization task.
% To extend the single domain task, we should split the single domain into pseudo multiple domain and perform the algorithm in future work.
% Beside, our method focus on the sub-problem of the domain bias, \textit{i.e.,} style bias.
% In future work, we should continue to explore broader solutions for DG-ReID.
% % Our method adopt meta-learning as our training scheme, which requires multiple source domains. 
% % In case there is only single source domain, we cannot directly apply our method to it. But we can split this domain into sub-domains by identity or camera and then applying our method on it. To validate the effectiveness of our method on single domain DG task is one of our future works.
\section{Conclusions}
\label{sec:conclusion}
% \lorem

In this paper, we propose a Style Variable and Irrelevant Learning (SVIL) method to eliminate the influence of style factors on the model, thus obtaining a more generalized model. 
More concretely, we design a Style Random Jitter (SJM) module to enrich the style diversity of source domains.
It can prompt the model focus on the identity-relevant information and ignore the style-relevant information.
Besides, we organically combine the SJM module with a meta learning algorithm, where the generalization ability of model is improve again.
% In which Style Random Jitter Module was proposed to mix the styles of source training domains to make the model learn style-invariant feature. Besides, we utilize MAML algorithm to collaborate with SJM so that the model can avoid overfitting to source domain bias. 
Extensive experiments demonstrate the effectiveness and superiority of our method.

% In this paper, we proposed a novel MAML-based style-invariant learning framework, named Style Variable and Irrelevant Learning (SVIL). In which Style Random Jitter Module was proposed to mix the styles of source training domains to make the model learn style-invariant feature. Besides, we utilize MAML algorithm to collaborate with SJM so that the model can avoid overfitting to source domain bias. Extensive experiments on large-scale DG ReID benchmarks show our SVIL can significantly outperforms state-of-the-arts, which demonstrate the effectiveness and superiority of our method.

\bibliographystyle{IEEEtran}
\bibliography{ref}
% \newpage

% \section{Biography Section}
% If you have an EPS/PDF photo (graphicx package needed), extra braces are
%  needed around the contents of the optional argument to biography to prevent
%  the LaTeX parser from getting confused when it sees the complicated
%  $\backslash${\tt{includegraphics}} command within an optional argument. (You can create
%  your own custom macro containing the $\backslash${\tt{includegraphics}} command to make things
%  simpler here.)
 
% \vspace{11pt}

% \bf{If you include a photo:}\vspace{-33pt}
% \begin{IEEEbiography}[{\includegraphics[width=1in,height=1.25in,clip,keepaspectratio]{fig1}}]{Michael Shell}
% Use $\backslash${\tt{begin\{IEEEbiography\}}} and then for the 1st argument use $\backslash${\tt{includegraphics}} to declare and link the author photo.
% Use the author name as the 3rd argument followed by the biography text.
% \end{IEEEbiography}

% \vspace{11pt}

% \bf{If you will not include a photo:}\vspace{-33pt}
% \begin{IEEEbiographynophoto}{John Doe}
% Use $\backslash${\tt{begin\{IEEEbiographynophoto\}}} and the author name as the argument followed by the biography text.
% \end{IEEEbiographynophoto}

\end{document}